%% file: example.tex
\documentclass{article}
\usepackage{amssymb}

\usepackage[preprint]{corl_2025} % Uncomment for pre-prints (e.g., arxiv); This is like ``final'', but will remove the CORL footnote.
\usepackage{times}

\usepackage[numbers]{natbib}
\usepackage{multicol}
\usepackage{wrapfig,booktabs}
\usepackage{multirow}
\usepackage{mathptmx}
\usepackage{graphicx}   % For including figures
\usepackage{caption}    % For caption formatting
\usepackage{float}      % For better float placement
\usepackage{stfloats}   % For two-column floats at the bottom of a page (optional)
\usepackage{silence}
\usepackage{amsfonts}
\usepackage[numbers]{natbib}
\usepackage[dvipsnames]{xcolor}
\input{sec/preamble}

\definecolor{darkpink}{RGB}{200, 75, 130} % Adjust RGB values for desired shade

\title{The One RING\includegraphics[height=6mm]{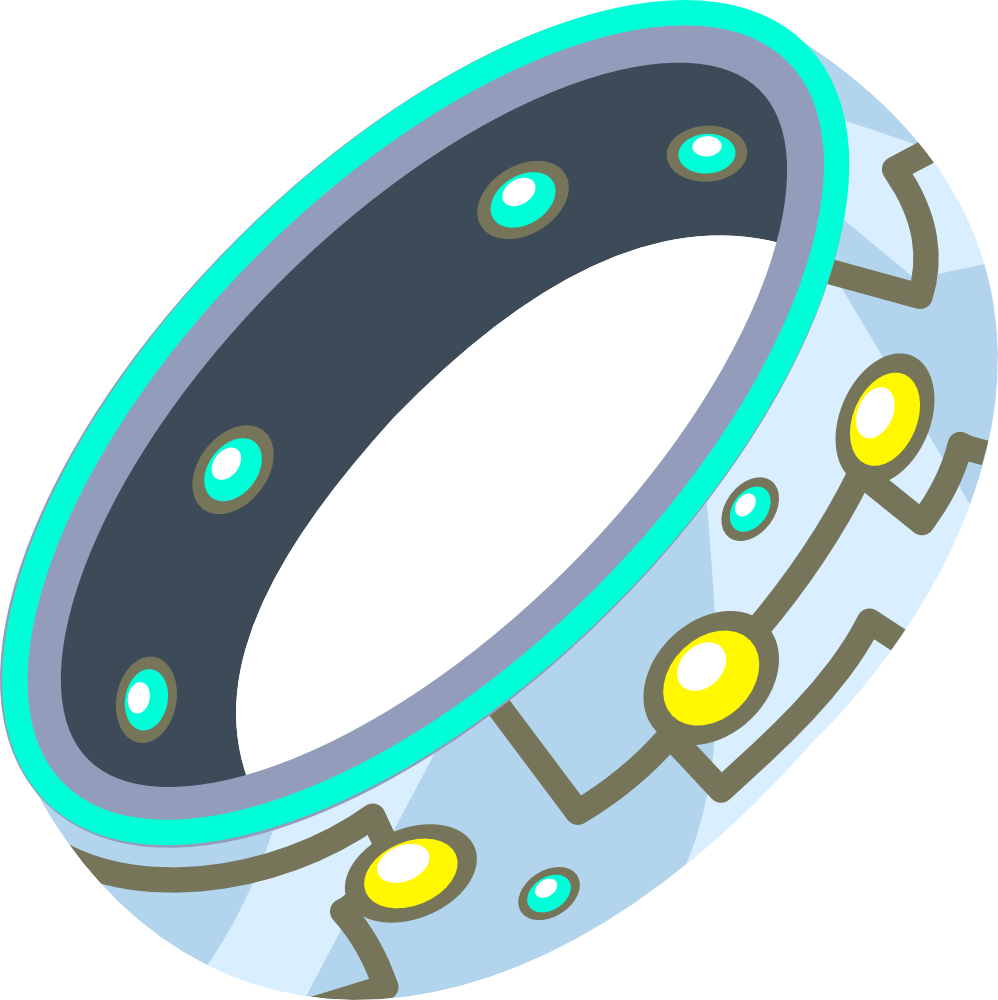}: \\a Robotic Indoor Navigation Generalist}

\author{ \hspace{0.5cm} \textbf{Ainaz Eftekhar }$^{1,2}$ \hspace{0.5cm} \textbf{Rose Hendrix}$^{\;2}$ \hspace{0.5cm} \textbf{Luca Weihs}$^{2}$ \hspace{0.5cm} \textbf{Jiafei Duan}$^{\;1,2}$ \hspace{1cm} \\ 
\hspace{0.5cm} \textbf{Ege Caglar}$^{1}$ \hspace{0.5cm} \textbf{Jordi Salvador}$^{\;2}$ \hspace{0.5cm} \textbf{Alvaro Herrasti}  $^{\;2}$ \hspace{0.5cm} \textbf{Winson Han}$^{2}$ \hspace{1cm} \\ \hspace{0.5cm} \textbf{Eli VanderBil}$^{2}$ \hspace{0.5cm} \textbf{Aniruddha Kembhavi}$^{\;1,2}$ \hspace{0.5cm} \textbf{Ali Farhadi} $^{\;1,2}$ \hspace{0.5cm} \textbf{Ranjay Krishna}$^{1,2}$ \hspace{1cm} \\ \textbf{Kiana Ehsani}$^{* 2}$ \hspace{1cm} \textbf{Kuo-Hao Zeng}$^{*2} $ \\ 
$^{1}$University of Washington\\
$^{2}$Allen Institute for AI
}

\begin{document}
\maketitle
\vspace{-0.4in}
\begin{center}
% \vspace{-4mm}
    \href{https://one-ring-policy.allen.ai/}{\textbf{\textcolor{darkpink}{one-ring-policy.allen.ai}}}
\end{center}

% \vspace{-4mm}
\begin{center}
    \centering 
    % \vspace{-0.175in}
    \includegraphics[width=\textwidth]{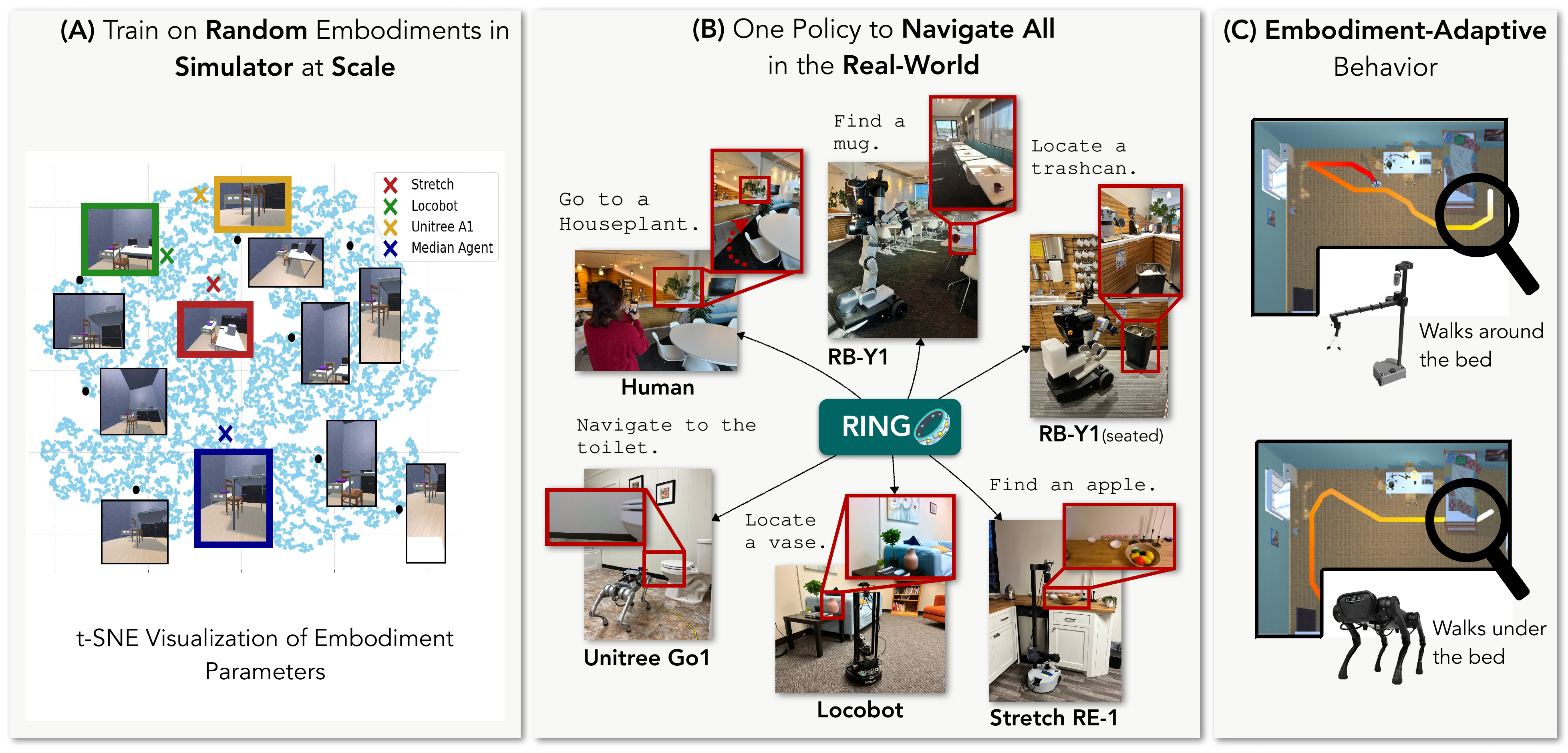}
    % \vspace{-4mm}
    \captionof{figure}{\footnotesize {We show that training on \textit{one million} randomly generated embodiments in simulation (varying camera configurations, body size, and rotation pivot point) results in \model, a generalist navigation policy that works across various robot embodiments in the real world. 
    (A) A t-SNE visualization of embodiment parameters 
    % parameters  $\mathbf{c}_e \in \mathbb{R}^{19}$ \
    for 30k random agents and three real robots (\textit{we do not train on any real robot embodiment parameters}). Egocentric views from the first camera are shown for 10 sample agents. 
    (B) \model transfers zero-shot to a wide range of embodiments in the real-world including Stretch RE-1, LoCoBot, Unitree Go1, RB-Y1 wheeled humanoid. % as well as human embodiment (navigation assitant).
    (C) The policy displays embodiment-adaptive behavior, adjusting its navigation strategy based on its embodiment.}}
    \label{fig:main_figure}
\end{center}

%===============================================================================

%===============================================================================

\input{sec/0_abstract}
\input{sec/1_intro}

\input{sec/2_related_work}
\input{sec/3_method}
\input{sec/4_results}

\input{sec/5_conclusion}
\input{sec/6_limitations}

%===============================================================================

% \section{Citations}
% \label{sec:citations}

% 	Citations can be made using either \textbackslash citep\{\} or \textbackslash citet\{\}, depending from the appropriateness. To avoid the citation moving to the next line, it is often a good practice to replace the space before with a tilde (\~{}) character.
% 	Example 1: ``CoRL is the best conference ever~\citep{Gauss1857}.''
% 	Example 2: ``\citet{Lagrange1788} proved, both theoretically and numerically, that CoRL is the best conference ever.''

%===============================================================================

\clearpage
% The acknowledgments are automatically included only in the final and preprint versions of the paper.
% \acknowledgments{If a paper is accepted, the final camera-ready version will (and probably should) include acknowledgments. All acknowledgments go at the end of the paper, including thanks to reviewers who gave useful comments, to colleagues who contributed to the ideas, and to funding agencies and corporate sponsors that provided financial support.}

%===============================================================================

\bibliography{example}  % .bib
\newpage
\input{sec/X_suppl}

\end{document}

%% file: sec/preamble.tex
\usepackage{graphicx}
\usepackage{caption}
\usepackage{subcaption}
\usepackage{multirow}
\usepackage{booktabs}
\usepackage{amsmath}
\usepackage{tablefootnote}
\usepackage{float}
\usepackage{xcolor}
\usepackage{amsmath}
\usepackage{pifont}
\usepackage{multirow}
\usepackage{xspace}
\usepackage[utf8]{inputenc}

\newcommand{\model}{\textsc{Ring}\xspace}

\newcommand{\fullname}{\textbf{\underline{R}}obotic \textbf{\underline{I}}ndoor \textbf{\underline{N}}avigation \textbf{\underline{G}}eneralist\xspace}

\newcommand{\modelzeroshot}{\textsc{Ring-Zero-Shot}\xspace}

\newcommand{\modellarge}{\textsc{Ring-Large}\xspace}

\newcommand{\modelembodimentconditioned}{\textsc{Ring-Emb-Cond}\xspace}

\newcommand{\cmark}{\ding{51}}  % Checkmark
\newcommand{\xmark}{\ding{55}}  % Crossmark

%% file: sec/0_abstract.tex
\begin{abstract}
Modern robots vary significantly in shape, size, and sensor configurations used to perceive and interact with their environments. However, most navigation policies are embodiment-specific—a policy trained on one robot typically fails to generalize to another, even with minor changes in body size or camera viewpoint. As custom hardware becomes increasingly common, there is a growing need for a single policy that generalizes across embodiments, eliminating the need to (re-)train for each specific robot. In this paper, we introduce \model (\fullname), an embodiment-agnostic policy that turns any mobile robot into an effective indoor semantic navigator. Trained entirely in simulation, \model leverages large-scale randomization over robot embodiments to enable robust generalization to many real-world platforms. To support this, we augment the AI2-THOR simulator to instantiate robots with controllable configurations, varying in body size, rotation pivot point, and camera parameters. On the visual object-goal navigation task, \model achieves strong cross-embodiment (XE) generalization—72.1\% average success rate across $5$ simulated embodiments (a 16.7\% absolute improvement on the \textsc{Chores}-$\mathbb{S}$ benchmark) and 78.9\% across $4$ real-world platforms, including Stretch RE-1, LoCoBot, and Unitree Go1—matching or even surpassing embodiment-specific policies. We further deploy \model on the RB-Y1 wheeled humanoid in a real-world kitchen environment, showcasing its out-of-the-box potential for mobile manipulation platforms.
% \keywords{Visual navigation, cross-embodiment, generalist navigation policy} 
\end{abstract}

%% file: sec/1_intro.tex
\section{Introduction}
\label{sec:intro}

Robot embodiments are diverse and are constantly evolving to better suit new environments and tasks.
%There is no clear consensus on key design choices: how many cameras a robot should have or whether it should move on two legs, four legs, or wheels. 
This range in body configurations—differences in size, shape, wheeled or legged locomotion, and sensor configurations—not only shapes how robots perceive the world but also how they act in it. A robot with a wide field of view (FoV) or multiple cameras can scan its surroundings quickly, while one with a narrower view might need to more actively explore a room. A small robot can squeeze through tight spaces, a low-profile one can duck under furniture, and a larger robot may need to follow more conservative routes. The influence of embodiment on behavior means a policy trained on one design, or even several, often does not perform well out of domain. 
% With designs changing so frequently, it is becoming increasingly important to develop \textit{transferrable policies} that can operate different robot types, eliminating the need for retraining \rose{awk revisit}. % covered in next para

There has been progress towards scalable cross-embodiment training~\citep{o2023open,team2024octo,doshi2024scaling, yang2024pushing,wang2024scaling}. While these methods demonstrate some transfer to unseen embodiments, they still suffer from performance degradation with relatively small changes in embodiment (e.g., camera pose modification on the same robot)~\cite{pumacay2024colosseum, xie2024decomposing}. Potentially, this is due to these methods relying on the small amount of real-world data available in public datasets-around 20 embodiments in total~\cite{o2023open}. Similarly, general-purpose navigation policies~\citep{shah2023gnm, shah2023vint, sridhar2023nomad} are trained on datasets with relatively few embodiments (e.g., 8 robots in~\citep{shah2023vint}), limiting their generalization. A more comprehensive solution is needed—one that can robustly handle the full spectrum of possible embodiments without retraining or additional adaptation. 

% Moreover, existing general navigation policies depend on constructing topological maps or graphs, while we focus on an end-to-end policy capable of exploring dynamic, unseen environments.

% There has been progress towards scalable cross-embodiment training~\citep{o2023open,team2024octo,doshi2024scaling, yang2024pushing,wang2024scaling} and in developing general-purpose navigation policies~\citep{shah2023gnm,shah2023vint,sridhar2023nomad}.  While these methods demonstrate some transfer to unseen embodiments, they require construction of topological maps or graphs and suffer performance degradation with relatively small changes in embodiment (e.g., camera position modification on the same robot). Potentially, this is due to these methods relying on the small amount of real-world data available in public datasets-only around 22 embodiments in total. We need a more comprehensive solution that reliably covers the wide range of possible embodiments without retraining or additional adaptation.

We introduce \model, a \fullname.
\model is trained \textit{exclusively in simulation}, without any use of real-world robot embodiments. In other words, all robot platforms we evaluate on (i.e., Stretch RE-1, LoCoBot, Unitree’s Go1, RB-Y1) are \textit{unseen} by \model during training.  We leverage simulation to randomly sample \textbf{1 Million} agent body configurations, varying the robot's camera parameters, collider sizes, and center of rotation.
Concretely, each embodiment consists of a collider box of varying dimensions and cameras with randomized parameters, placed randomly within the collider box. Fig.\ref{fig:main_figure}-A presents a t-SNE\citep{van2008visualizing} visualization of body parameters for 30k such agents. Our approach builds on the success of prior works that achieve strong real-world performance through large-scale simulation-only training~\citep{ehsani2024spoc,zeng2024poliformer,hu2024flare}. Simulation enables training across a vast distribution of environments (150k ProcTHOR houses~\citep{Deitke2022ProcTHORLE}) and objects (40k+ 3D objects in Objaverse~\citep{Deitke2022ObjaverseAU}) in the AI2-THOR simulator. Extensive domain randomization on visual observations and the use of pre-trained visual encoders then allows simulation-trained policies to bridge the sim-to-real gap. We follow the training procedure in FLaRe~\citep{hu2024flare}, first training our policy on expert trajectories collected from $1$M randomized embodiments and subsequently fine-tuning it with on-policy reinforcement learning (RL) in the simulator.

Our results demonstrate generalization to \textit{truly unseen embodiments}. \model transfers to diverse real-world embodiments without any adaptation, despite being trained entirely in simulation without access to the real robot configurations. We evaluate in a zero-shot setting across Stretch RE-1, LoCoBot, Unitree’s Go1, RB-Y1 wheeled humanoid, and even ``Navigation Assistants,'' where a human user captures ego-centric observations on their phone and prompts \model to predict actions. \model achieves $72.1\%$ average success rate in simulation (16.7\% absolute improvement on \textsc{Chores}-$\mathbb{S}$ benchmark)  and $78.9\%$ on real robot platforms—matching or even surpassing embodiment-specific policies.
% \model achieves $72.1\%$ and $78.9\%$ average success rates, significantly outperforming the best baseline in both simulation and the real world.
\model can be further adapted to an \textit{embodiment-specialized} policy with even better performance (up to 10\% absolute improvement) with minimal finetuning.
\model is easy to install, and is ready for use by the community. We will release our pretrained models, code, and data.

%% file: sec/2_related_work.tex
\section{Related work}
% \vspace{-2mm}

\noindent\textbf{Cross-embodiment.}
Cross-embodiment training has received substantial attention from the research community.
Arguably the most representative of a large body of recent work~\cite{yang2024pushing,doshi2024scaling,octo_2023,wang2024scaling,duan2024manipulate,loquercio2018dronet,xu2024flow,ha2024umi,hejna2024re,chen2024rovi,harithas2024motionglot,bharadhwaj2024track2act,kedia2024one,kim2024openvla,xu2023xskill,zakka2022xirl}, Open-X-Embodiment (OXE) \cite{open_x_embodiment_rt_x_2023} is the fruit of a large collaboration to cover many robotic tasks, with special emphasis in manipulation. Its usage in RT-X results in a notable performance gain in emergent skill evaluations in comparison to RT-2 \cite{Brohan2023RT2}. Despite the 1.5 million trajectories across 22 embodiments present in their dataset, the enormous cost of data collection in the real world makes further scaling challenging. CrossFormer~\cite{doshi2024scaling} trains a transformer-based policy on 900k trajectories spanning 30 robots, drawing from OXE, navigation data from GNM~\cite{shah2023gnm}, manipulation data from DROID~\cite{khazatsky2024droid}, and additional sources. However, the limited diversity of embodiments and focus on low-level control highlight the need for denser embodiment coverage.
GET-zero~\cite{patel2024getzero}, focused on dexterous manipulation, incorporates embodiment structure via a connectivity graph to guide attention.
In contrast, we generate an arbitrarily large set of randomized embodiments during training, allowing our policy to generalize zero-shot to novel embodiments without requiring access to their structure.
% CrossFormer~\cite{doshi2024scaling} trains a transformer-based policy on 900k trajectories across 30 robots, including a subset of OXE, navigation data from GNM \cite{shah2023gnm}, manipulation data from DROID \cite{khazatsky2024droid}, and additional collected data. Due to the relatively sparse amount of embodiments observed during training and the target low-level control, it showcases the need for denser embodiment coverage.
% GET-zero \cite{patel2024getzero} focuses on dexterous manipulation, and proposes to inform the policy with the structure of the embodiment via a connectivity graph to bias the attention.
% In contrast, we generate an arbitrarily large amount of embodiments for training our policy, enabling zero-shot deployment to new embodiments without accessing the embodiment structure.

\noindent\textbf{Foundational navigation policies}
Following the success in recent developments for point-goal navigation~\cite{Wijmans2020DDPPO}, locomotion~\cite{radosavovic2024real,bohlingerone,shafiee2024manyquadrupeds, Ying2024CEURL}, agile control~\cite{xiao2024anycar}, exploration~\cite{chaplot2020learning,yamauchi1997frontier,ye2021auxiliary}, and social navigation~\cite{Puig2023Habitat3}, comparable results in more nuanced tasks like semantic or object-goal navigation (ObjectNav)~\cite{Batra2020ObjectNav,eftekhar2023selective,zeng2021pushing,majumdar2022zson,wani2020multion,Ramrakhya2022HabitatWeb,yokoyama2024hm3d,khanna2024goat} remain elusive due to a lack of efficient exploration and semantic understanding capabilities.
Recently, with powerful pretrained vision models~\cite{oquab2023dinov2,Zhai2023SigLIP} and large-scale procedurally generated virtual environments \cite{Deitke2022ObjaverseAU}, notable progress in end-to-end ObjectNav policy learning \emph{for specific embodiments} has been achieved by means of imitation learning (IL) from shortest-path trajectories~\cite{ehsani2024spoc}, RL~\cite{zeng2024poliformer}, or combinations thereof~\cite{hu2024flare}.
In image-goal navigation, NoMaD \cite{sridhar2023nomad}, which extends ViNT \cite{shah2023vint}, uses a diffusion policy to control a single embodiment.
With the same goal in mind, GNM \cite{shah2023gnm} trains navigation policies across 6 embodiments using IL. 
In contrast, our policy benefits from RL fine-tuning, improving robustness to compounding errors.
Leveraging large-scale training with randomized embodiments in simulation, \model learns a single policy that generalizes to \emph{any} embodiment, including \emph{truly} unseen robot platforms in the real world. Unlike NoMaD, ViNT, GNM, and Mobility VLA~\cite{xumobility}, which rely on topological map or graph reconstruction for high-level planning, our approach is fully end-to-end and can explore dynamic novel scenes without requiring an explicit map. While prior work~\cite{nasiriany2024pivot, yuan2024robopoint, ahn2022can} has explored embodiment-agnostic policies using LLMs or VLMs, these methods are limited to short-horizon navigation and single-step prediction. In contrast, \model incorporates temporal context via a transformer decoder.
% In contrast, our policy benefits from finetuning with RL, improving resilience to compounding errors.
% Additionally, thanks to training with large-scale randomized embodiments in simulation, \model learns a single policy to navigate \emph{any} embodiment, generalizing to \emph{truly} unseen robot platforms in the real world.
% Furthermore, NoMaD, ViNT, GNM, and Mobility VLA~\cite{xumobility} all require topological map or graph reconstruction for high-level planning, whereas our policy is fully end-to-end and explores novel scenes without an explicit map.
% While several efforts~\cite{nasiriany2024pivot, yuan2024robopoint,ahn2022can} focus on learning embodiment-agnostic policies using LLMs or VLMs, they  address only short-horizon navigation tasks and perform single-step predictions. In contrast, \model models temporal information through a transformer decoder.

%% file: sec/3_method.tex
\section{\model}
% \vspace{-2mm}
% While recent advances in robotic learning have enabled the training of successful navigation policies, these policies often overfit to a specific robot platform, struggling to generalize beyond the trained embodiments. 

With the growing diversity of robots used in research labs and real-world applications, there remains a need for a policy that can operate a wide range of embodiments and transfer, in a zero- or few-shot manner, to unseen robots. 
We introduce \model, a generalist policy for indoor visual navigation that \textit{learns from a broad spectrum of embodiments, trained exclusively in simulation, without any direct use of actual robot embodiments}. We show that training on an extensive range of ${\sim}1$M random embodiments results in a robust navigation policy, enabling zero-shot transfer to unseen real-world robots.
To train \model, we define the space of random embodiments (Sec. \ref{sec:emb_random}), enable generation of expert trajectories for random embodiments in simulation (see Appendix \ref{app:experts}), and use state-of-the-art architecture designs (Sec. \ref{sec:architecture}) to train with a combination of IL and RL methods (Sec. \ref{sec:training}).

%\subsection{Cross-Embodiment Policy: Problem Formulation}

\subsection{Problem formulation}

We define the space of possible embodiments as $E$, where each embodiment $e \in E$ is characterized by a configuration vector $\mathbf{c}_e$, including parameters such as camera settings, agent collider size, and center of rotation. Each task can be modeled as a Partially Observable Markov Decision Process (POMDP), denoted as $(S, A, E, O_e, T_e, R, L, P(s_0), \gamma)$, where $S$ and $A$ are the state and action spaces. The observation space $O_e$ varies across embodiments due to differences in camera parameters.
The observation at time $t$ for embodiment $e$, $o_t^e = O_e(s_t, \mathbf{c}_e)$, is a function of both the state $s_t$ and embodiment parameters $\mathbf{c}_e$. Given an action $a_t$, the next state follows the transition dynamics $s_{t+1} \sim T_e(s_{t+1} | s_t, a_t, \mathbf{c}_e)$, which depends on the embodiment (due to variations in collider size and rotation center). Fig.~\ref{fig:2_embodiments} shows trajectories from two different embodiments starting at the same location and following the same sequence of actions. They have distinct visual observations and follow different transition dynamics—one agent moves under the table, while the other collides with it.

Except where otherwise specified, we assume that all embodiments share the same discrete action space \{\texttt{MoveBase(±20cm)}, \texttt{RotateBase(±$6^\circ$, ±$30^\circ$)}, \texttt{Done}\}. These actions are executed using robot-specific low-level controllers during deployment. This simple and platform-agnostic action space enables effective cross-embodiment (XE) transfer, as demonstrated across both holonomic and differential-drive robots in our experiments. Investigating more expressive XE action spaces beyond this sufficient version is left for future work.

\input{tables/body_randomization}

\subsection{Embodiment randomization at scale}
\label{sec:emb_random}

Domain randomization~\cite{Chen2022DomainRand} is a class of methods in which policies are trained across a wide range of simulated \textit{environmental} parameters; the aim is to enable robustness to unseen environments.
Our approach is complementary yet orthogonal; we apply embodiment randomization to train policies on a diverse set of \textit{robot body} parameters, enabling robust deployment to unseen real-world robots.
\input{figs/2_embodiments}

We model the body of the agent as an invisible collider box in the AI2-THOR~\cite{Kolve2017AI2THORAI} simulator. Each agent can have 1 or 2 RGB cameras placed at a random pose within the collider box. Parameters corresponding to both the body and the cameras are sampled randomly from the ranges specified in Tab.~\ref{tab:body_params}. We also modify the process of generating expert trajectories to account for the diversity of embodiments, for details see Appendix~\ref{app:experts}. Below, we detail the parameters varied in our embodiment randomization: \textbf{Collider size $(\alpha_x, \alpha_y, \alpha_z)$}. The agent's body is modeled as a collider box. We use three scale factors $(\alpha_x, \alpha_y, \alpha_z)$ to scale the box along $x, y, z$ axis. \textbf{Rotation center ($o_x, o_y, o_z$)}. These coordinates define the agent’s pivot point. While this center is typically near (0,0), it can vary across different robots. We sample $o_x$ from the range $[-\frac{\alpha_x}{3}, \frac{\alpha_x}{3}]$ and $o_y$ from the range $[-\frac{\alpha_y}{3}, \frac{\alpha_y}{3}]$, with the sampling ranges determined by the collider size. \textbf{Camera parameters}. Each agent is equipped with two RGB cameras placed within the collider box. We randomize several camera parameters, including position, rotation, FoV, and aspect ratio. While the first camera always faces forward, the second camera can rotate up to $360^{\circ}$ in $z$-axis, enabling it to face forward, to the sides, or backward.

% \kiana{the following paragraph with its figure can be moved to supp imo}

% For visualization purposes, we define an embodiment configuration vector $\mathbf{c}_e \in \mathbb{R}^{19}$ for each embodiment, representing the camera and body parameters. Fig.~\ref{fig:main_figure}-A shows a t-SNE visualization of the vectors $\mathbf{c}_e$ for 30k of our random embodiments, along with the corresponding vectors for Stretch RE-1, LoCoBot, and Unitree A1. The figure also includes the egocentric view from the first camera for 10 random embodiments, and the three robots. This demonstrates that our randomization spans a wide range of possible embodiments, covering the real-world robot platforms of interest. 
% (more details in Appendix~\ref{app:tsne_nn}). 
\input{figs/model}

\subsection{Architecture}
\label{sec:architecture}

% ORIGINAL
% As each timestep, the input to \model includes one or two visual observations captured from the cameras and language instruction $l$. \model then predicts action probability over the action space and sample an action to execute.
% \model's architecture is inspired by PoliFormer~\cite{zeng2024poliformer} and consists of a Visual Encoder, a Goal Encoder, a Transformer State Encoder, and a Causal Transformer Decoder with a linear actor-critic head. The Visual Encoder is a visual transformer that encodes two RGB observations from the two cameras into visual token features; the Goal Encoder is a language model that encodes the language instruction $l$ into goal token features. We then concatenate the visual token features, goal token features, and a special \texttt{STATE} token feature along the token dimension. The Transformer State Encoder summarizes the state at each timestep by performing layers of self-attention across the concatenated token features. It returns the hidden feature corresponding to the \texttt{STATE} token. Finally, the Causal Transformer Decoder performs explicit memory modeling over time. It applies layers of causal self-attention on the hidden features returned from State Encoder along the temporal axis. The linear actor-critic head further predicts action logits over the action space as well as a value estimation at the latest timestep. Please find more details, such as the input and hidden dimensions, number of layers of each module, and a figure illustrating our policy model, in App.~\ref{sec:ablations_agent_param}.

With this rich dataset of expert trajectories for random embodiments, a deep, high-capacity architecture is essential to learn a robust policy (Fig.~\ref{fig:model}). At each timestep, \model uses $N$ RGB images (one per camera) and a language instruction $l$ to predict an action distribution over a discrete action space. To account for different dimensions, we pad the RGB observations to square and resize to $256 \times 256$ before feeding them to the model.
\model's architecture, inspired by PoliFormer~\cite{zeng2024poliformer}, consists of a Visual Encoder, a Goal Encoder, a Transformer State Encoder, and a Causal Transformer Decoder with a linear actor-critic head. The Visual and Goal Encoders are frozen pre-trained models (a ViT and a language model, respectively) that encode the RGB observations and instruction $l$ into visual and goal token embeddings. Projections of these embeddings, along with a special \texttt{STATE} token vector, are stacked along the token axis and processed by the multi-layer Transformer State Encoder, which summarizes the observations at each timestep as the state embedding corresponding to the \texttt{STATE} token. Finally, the Causal Transformer Decoder performs explicit memory modeling over time, producing the current belief by causal attention on the state embeddings stacked along the temporal axis. The linear actor-critic head further predicts action logits over the action space as well as a value estimate. (More details in Appendix~\ref{app:model_arch_details})

\subsection{Training paradigm}
\label{sec:training}

% Recently, SPOC~\cite{ehsani2024spoc} showed that training policies with Behavior Cloning on large-scale expert trajectories in simulation leads to policies that effectively generalize to the real world. FLaRe~\cite{hu2024flare} further introduced a robust and scalable method for finetuning such pretrained policies with On-policy Reinforcement Learning. RL finetuning introduces error recovery behaviors and mitigates the compounding errors typically encountered in imitation learning, leading to a substantial performance boost. 

We adopt the training recipe of pretraining our policy on expert trajectories collected from randomized embodiments (Sec.~\ref{sec:emb_random}), followed by finetuning with on-policy RL using the randomized embodiments in the AI2-THOR simulator~\cite{Kolve2017AI2THORAI}.

% \kiana{this next paragraph is super repeated! only 1M traj from 50k houses is new. }
% \noindent\textbf{Large-scale imitation learning with random embodiments.} We collect a large-scale dataset of expert trajectories using A* planners within the AI2-THOR simulator~\cite{Kolve2017AI2THORAI}. For each trajectory, we sample an embodiment from the ranges specified in Table~\ref{tab:body_params} and select a scene from one of the Objaverse-Populated ProcTHOR houses~\cite{Deitke2022ObjaverseAU, Deitke2022ProcTHORLE}. The agent navigates to the target object using an A* algorithm-based planner. In total, we gather \textbf{1M} trajectories across 50K houses, each with a randomly sampled embodiment. Fig.~\ref{fig:2_embodiments} shows sample trajectories from two random agents.

\noindent\textbf{Large-scale imitation learning with random embodiments:}
We train our policy using expert trajectories collected from 1M randomized embodiments across 50k procedurally generated \textsc{ProcTHOR} houses~\cite{Deitke2022ProcTHORLE}, containing approximately 40k annotated 3D objects~\cite{objathor2024}. At each time step, the linear actor-critic head in the Causal Transformer Decoder predicts action logits, and a cross-entropy loss is computed between the predicted logits $\pi^t$ and the expert action. We use a batch size of 240 trajectories, each with a temporal context window of 100 steps. Training is conducted on 8$\times$ H100 GPUs (80 GB each) using the AdamW optimizer with a learning rate of $2 \cdot 10^{-4}$ for 80k iterations.

\noindent\textbf{Large-scale RL finetuning with random embodiments:} Following the training recipe in FLaRe~\cite{hu2024flare}, we perform large-scale RL fine-tuning using AllenAct~\cite{AllenAct} on randomized embodiments in simulation. This fine-tuning phase is critical for enabling the policy to learn through trial and error how to navigate diverse embodiments. We use DD-PPO with 64 parallel environments and 128 rollout steps across 4 machines (each with 8$\times$ H100 GPUs), training for 40M steps using the AdamW optimizer with a learning rate of $2{\cdot}10^{-5}$. As in FLaRe~\cite{hu2024flare}, we disable the entropy term in the PPO loss to prevent catastrophic forgetting. For fair comparison, we adopt the reward function from~\cite{Deitke2022ProcTHORLE}: $r_t = \max\left(0, min\Delta_{0:t-1} - \Delta_t\right) + s_t - \rho$, where $\min\Delta_{0:t-1}$ is the minimum L2 distance between the agent and the target object up to time $t{-}1$, $\Delta_t$ is the current L2 distance, $s_t$ is a success reward, and $\rho = 0.01$ is a step penalty encouraging task efficiency. The agent must explicitly issue \texttt{Done} to receive the success reward ($s_t=10$); otherwise, $s_t=0$.

%% file: tables/body_randomization.tex
\begin{wraptable}{R}{0.48\textwidth}
\vspace{-10mm}
\centering
\scriptsize
\addtolength{\tabcolsep}{-1.5em}
\begin{tabular}{lc}
\hline Parameters & Training Range \\
\hline 
Collider Size ($\alpha_x, \alpha_y, \alpha_z$) & {$[0.2, 0.5]$, $[0.3, 1.5]$, $[0.2, 0.5]$} \\
Rotation Center ($o_x, o_y, o_z$) & {$[-\alpha_x/2, \alpha_x/2]$, $[-\alpha_y/2, \alpha_y/2]$, $[-\alpha_z/2, \alpha_z/2]$} \\
Vertical FoV (${\text{cam1}}, {\text{cam2}}$) & {$[40, 100]$, $[40, 100]$} \\
Horizontal FoV (${\text{cam1}}, {\text{cam2}}$) & {$[40, 120]$, $[40, 120]$} \\
Camera Pitch (${\text{cam1}}$) & {$[-20, 40]$} \\
Camera Pitch (${\text{cam2}}$) & {$[-20, 60]$} \\
Camera Yaw (${\text{cam1}}, {\text{cam2}}$) & {always 0, $[0, 360]$} \\
Camera Position (x) (${\text{cam1}}, {\text{cam2}}$) & {$[-\alpha_x/2, \alpha_x/2]$, $[-\alpha_x/2, \alpha_x/2]$} \\
Camera Position (y) (${\text{cam1}}, {\text{cam2}}$) & {$[0.3, \alpha_y]$, $[0.3, \alpha_y]$} \\
Camera Position (z) (${\text{cam1}}, {\text{cam2}}$) & {$[-\alpha_z/2, \alpha_z/2]$, $[-\alpha_z/2, \alpha_z/2]$} \\
RGB dimensions (H, W) & {$[112, 448]$, $[112, 448]$} \\
\hline
\end{tabular}%
\caption{\footnotesize{\textbf{Random Embodiment Parameters.} 1M embodiments are sampled from these ranges.}}
\label{tab:body_params}
% \vspace{-1mm}
\end{wraptable}

%% file: figs/2_embodiments.tex
% \begin{figure}[t]
% \vspace{-1mm}
% %\hspace{4mm}
% \centering
% \includegraphics[width=1\columnwidth]{figs/2_embodiments.pdf} 
% \caption{
% \textbf{Different embodiments exhibit different behaviors.} 
% For each embodiment in these sample trajectories, the left column shows the first-person view from the main camera and the second one a third-person view of the agent --white boxes indicate the robot colliders. Embodiment A (shown on the left) has a bigger body size compared to Embodiment B (shown on the right). As a result, B can go under the table to get to the chair but A collides with the table and has to go around.
% % For each embodiment, the left column shows the first-person view from the main camera and the second one a third-person view of the agent.
% % The white boxes indicate the robot colliders.
% % Both agents start from the same position and execute the same sequence of actions. Visual observations and transition dynamics differ notably depending on the embodiment. For instance, the shorter agent (Embodiment B) can move under the table, while the taller one (Embodiment A) collides with it.
% }
% \label{fig:2_embodiments}
% \end{figure}

\begin{wrapfigure}{R}{0.48\textwidth}
\vspace{-4mm}
  \begin{center}
    \includegraphics[width=0.51\textwidth]{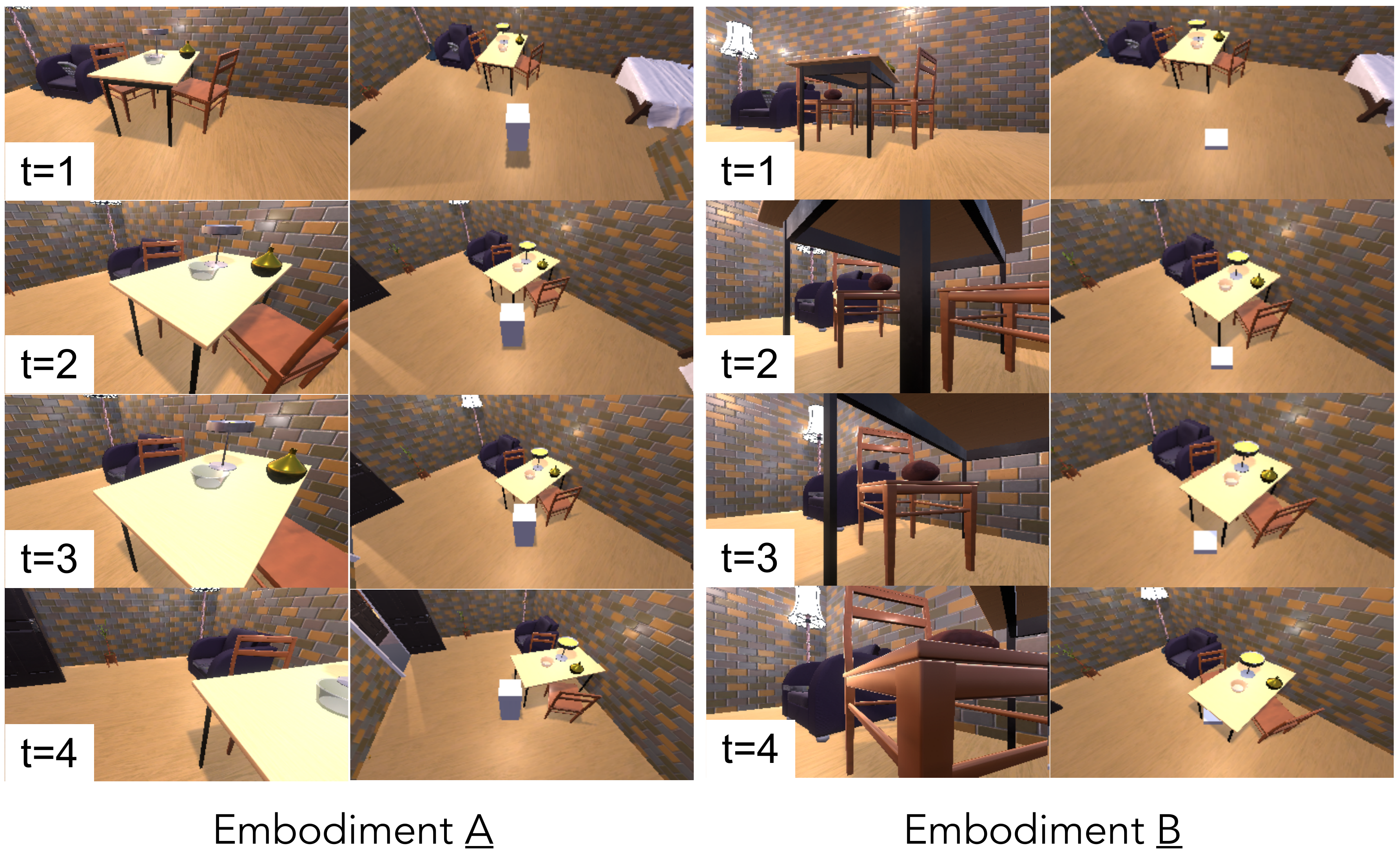}
  \end{center}
  \vspace{-4mm}
  \caption{\footnotesize{\textbf{Different embodiments exhibit different behaviors.} 
We show egocentric view from the main camera and third-person view of the 2 agents--white boxes indicate the robot colliders. Embodiment B can go under the table to get to the chair but Embodiment A collides with the table and has to go around.}}
\label{fig:2_embodiments}
  \vspace{-4mm}
\end{wrapfigure}

%% file: figs/model.tex
% \begin{figure}[H]
% 	\centering
% 	\includegraphics[width=1\columnwidth]{figs/model.pdf}
% 	\caption{\textbf{Our \model model architecture}. It accepts visual observations and a language instruction as inputs and predicts an action to execute. At RL finetuning, \model also predicts a value estimate.}
% 	\label{fig:model}
% \end{figure}

\begin{wrapfigure}{R}{0.46\textwidth}
\vspace{-12mm}
  \begin{center}
    \includegraphics[width=0.46\textwidth]{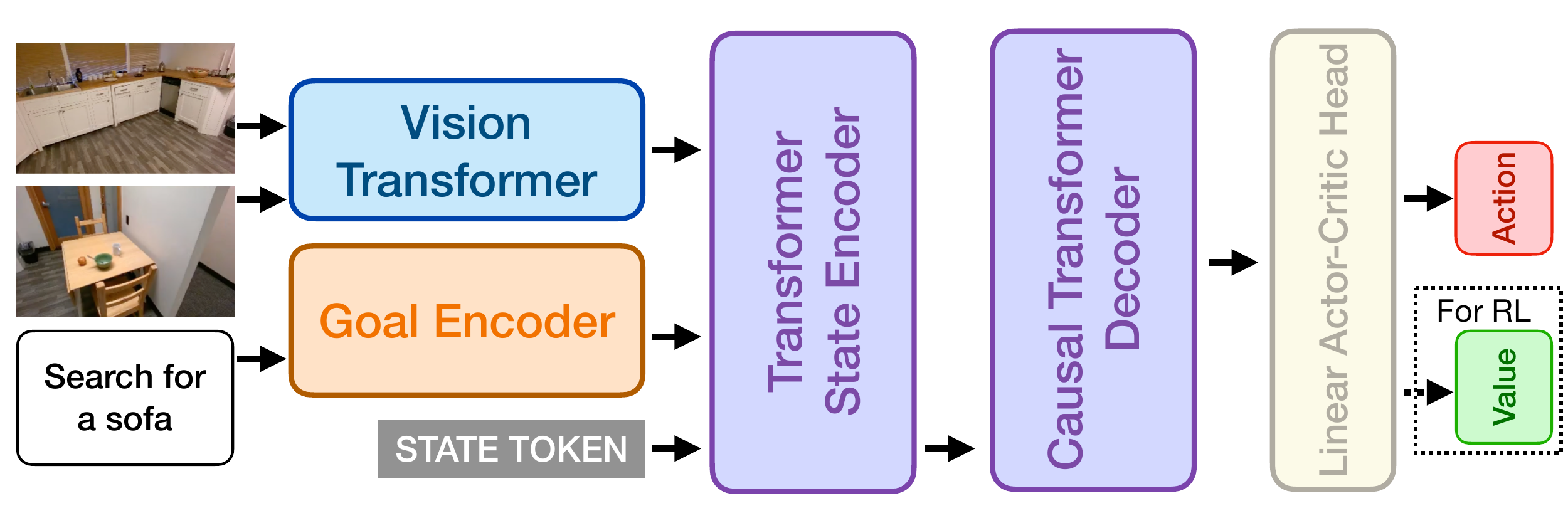}
  \end{center}
  \vspace{-4mm}
  \caption{\footnotesize{\textbf{\model model architecture}. It accepts visual observations and a language instruction as inputs and predicts an action to execute. At RL finetuning, \model also predicts a value estimate.}}
	\label{fig:model}
  \vspace{-4mm}
\end{wrapfigure}

%% file: sec/4_results.tex
\section{Experiments}
\input{tables/zero_shot}
% \vspace{-2mm}

Our experiments show that \model operates effectively across a wide range of embodiments, including \textit{Stretch RE-1}, \textit{LoCoBot}, \textit{Unitree Go1}, and \textit{RB-Y1}, despite being trained exclusively in simulation \textbf{without} any direct exposure to real robot embodiments.
Our key results are: 1)  \model generalizes \textbf{zero-shot} to 5 \textit{truly} unseen embodiments, despite never being trained on them, and achieves state-of-the-art performance across multiple benchmarks (Sec.~\ref{sec:zero_shot}). 2) Our policy, trained solely in simulation on randomized embodiments, transfers directly to the \textbf{real-world}, on 5 real robots and as navigation assistants (Sec.~\ref{sec:real_world}). \textit{(Videos in supplementary)} 3) \model can be easily adapted to \textbf{embodiment-specialized} policies with minimal finetuning. It achieves better performance on each specific robot (Sec.~\ref{sec:finetuning}). 4) \model shows \textbf{embodiment-adaptive behavior}, adjusting its strategies based on the agent's body (Sec.~\ref{sec:emb_aware}). 5) We perform a \textbf{collision analysis} showing that \model remains as safe—\textit{and in some cases even safer}—than embodiment-specific policies(Sec.~\ref{sec:ablations}).

% \begin{enumerate}
% \item \model generalizes \textbf{zero-shot} to 4 \textit{truly} unseen embodiments, despite never being trained on them, and achieves state-of-the-art performance across multiple benchmarks (Sec.~\ref{sec:zero_shot}). 
% \item Our policy, trained solely in simulation on randomized embodiments, transfers directly to the \textbf{real-world}, on 4 real robots and navigation assistants (human evaluation) (Sec.~\ref{sec:real_world}). \textit{(Qualitative videos available in the supplementary material)}
% \item \model can be easily adapted to \textbf{embodiment-specialized} policies with minimal finetuning. It achieves better performance on each specific robot (Sec.~\ref{sec:finetuning}).
% \item \model shows \textbf{embodiment-adaptive behavior}, adjusting its strategies based on the agent's body (Sec.~\ref{sec:emb_aware}). 
% \item We present ablation studies and explore finetuning with collision penalties to enable the policy to take more conservative actions (Sec.~\ref{sec:ablations}). 
% \end{enumerate}

%\kiana{definitely need to compare with one of dhruv shah's work as baseline. or train a model that is somehow similar to that. }

\subsection{\textbf{\model generalizes zero-shot to unseen embodiments}}
\label{sec:zero_shot}

We perform zero-shot evaluations of all policies on four robot embodiments: Stretch RE-1 (with 1 or 2 cameras), LoCoBot, and Unitree A1 in simulation.

\noindent \textbf{Baselines.} We select prior works from both imitation learning (IL) and reinforcement learning (RL) for comparison. Each baseline is trained on a specific embodiment and evaluated in a zero-shot setting across four different embodiments. SPOC~\cite{ehsani2023imitating} is a supervised IL baseline trained on shortest-path expert trajectories in AI2-THOR. PoliFormer~\cite{zeng2024poliformer} is a state-of-the-art transformer-based policy for object goal navigation, trained from scratch using RL. FLaRe~\cite{hu2024flare} combines IL and RL for efficient policy fine-tuning. All baselines use similar architectures and comparable data, except for SPOC (reason to include SPOC 2.3M). Specifically, SPOC~\cite{ehsani2024spoc} (SPOC-$2.3$M) is trained with IL on Stretch RE-1 using $100$k ($2.3$M) expert trajectories; Poliformer~\cite{zeng2024poliformer} is trained from scratch on each embodiment individually over $300$M RL steps (more than other baselines in terms of training frames); and FLaRe~\cite{hu2024flare} finetunes SPOC on Stretch RE-1 with an additional $20$M RL steps. 

\noindent\textbf{Experimental details.}
\model is first trained with IL on $1$M expert trajectories collected from randomized embodiments in simulation, followed by finetuning with RL for an additional $40$M steps on the randomized embodiments. \textit{Note that all four target embodiments were unseen during training}. We evaluate on the navigation benchmark in \textsc{Chores}-$\mathbb{S}$~\cite{ehsani2024spoc}, a simulation benchmark for household robot with 200 tasks across 200 scenes. For Unitree A1, we create a new, similar benchmark with 200 tasks adjusted for the robot's lower height to ensure that all targets are feasible. 

\noindent\textbf{Results.} Tab.~\ref{tab:zero_shot} presents the zero-shot evaluation of all policies across four embodiments. We compare \textit{Success Rate} and \textit{Success Weighted by Episode Length (SEL~\cite{eftekhar2023selective})}, a metric measuring efficiency. The results indicate that all single-embodiment baselines struggle to generalize effectively to new embodiments, with performance declining as embodiment differences increase. In contrast, \model exhibits strong generalization across all embodiments, despite not being trained on any of them, achieving an average absolute improvement of $16.7\%$ in Success Rate. In some cases, it outperforms the baseline trained on the target embodiment: PoliFormer trained on LoCoBot \textit{($61.5 \rightarrow 68.5$)} and Unitree A1 ($55.3 \rightarrow 72.0$). 
These 2 more challenging embodiments (lower FoV, low camera placement) make RL from scratch less effective. \model benefits from more efficient learning by training across random embodiments at scale with more diverse navigation behaviors.
% This shows that \model benefits from training across random embodiments at scale, leading to a more effective navigation policy which even outperforms some embodiment-specialized policies.

\subsection{\textbf{\model transfers to real-world embodiments despite being purely trained in simulation}}
\label{sec:real_world}

\textbf{Robot evaluation.} We zero-shot evaluate our policy on 4 unseen robots in a multi-room real-world apartment (Fig.\ref{fig:real_layout}) without any real-world-specific finetuning (Tab.~\ref{tab:real_world}). We use the same evaluation set of 15 tasks for LoCoBot\cite{Deitke2022ProcTHORLE,Deitke2023Phone2Proc,zeng2024poliformer} (3 start poses × 5 targets) and 18 tasks for Stretch RE-1~\cite{ehsani2024spoc,zeng2024poliformer,hu2024flare} (3 poses × 6 goals). For Unitree Go1, we create a new set with 3 start poses and 4 objects (\textit{toilet, sofa, TV, trashcan}) placed to match its lower viewpoint. \model matches or outperforms specialized policies, likely due to cross-embodiment (XE) training enabling robust sim-to-real transfer in the presence of real-world noise. We also deploy \model on the RB-Y1 wheeled humanoid in an unstructured kitchen, where it successfully navigates to targets (\textit{trashcan, apple, houseplant, mug}) at two different heights (standing/seated), using an iPhone 16 Pro camera mounted on the robot to stream visual observations (Fig.~\ref{fig:main_figure}-B).
(\textit{More detail in Appendix~\ref{app:real_robot_details} \& videos in supplementary.})

\input{tables/real_world}

\noindent\textbf{Human evaluation.}
To further show \model's generalization to unseen embodiments, we evaluate our policy as a navigation assistant with humans as novel embodiments. Five participants navigated a real-world kitchen by following policy outputs on their phones. Each had unique characteristics (e.g., step size, height, rotation, camera posture) and was tasked with reaching three objects (\texttt{Mug}, \texttt{Apple}, \texttt{Houseplant}), yielding 15 trajectories. We compare \model to FLaRe~\cite{hu2024flare}, trained only on Stretch RE-1. As shown in Tab.\ref{tab:humans}, \model consistently outperforms FLaRe across objects and users. Fig.\ref{fig:human_eval} shows two qualitative examples (more in Appendix~\ref{app:human_eval_details}).

\input{figs/bed_trajectory}
\subsection{\textbf{\model can efficiently adapt to an embodiment-specialized policy with minimal finetuning}}

\label{sec:finetuning}
\input{figs/finetune}
Although \model generalizes zero-shot across diverse embodiments, some scenarios benefit from embodiment-specialized policies for optimal performance. Here, we show that \model can be easily adapted to a \textit{robot-specialized policy} through minimal fine-tuning. 
\textbf{Baselines.} We compare with FLaRe~\cite{hu2024flare}, which demonstrates effective adaptation to new tasks and embodiments. It is pretrained on Stretch RE-1 and finetuned on each of the three test embodiments using up to $20$M RL steps. \textbf{Implementation.} We finetune \model, pretrained on randomized embodiments, on each robot for up to $20$M RL steps, using the same hyperparameters as FLaRe for fair comparison. Following FLaRe, we repurpose \texttt{RotateBase(±$6^\circ$)} to \texttt{TiltCamera(±$30^\circ$)} for LoCoBot, enabling camera control not available during zero-shot evaluation. \textbf{Results.} As shown in Fig.~\ref{fig:finetune}, \model adapts efficiently, achieving superior performance with minimal fine-tuning. For LoCoBot and Unitree-A1, FLaRe underperforms compared to Stretch RE-1, suggesting that pretraining on a single embodiment limits generalizability. This underscores the value of policies like \model that can adapt quickly and consistently to new embodiments.

\subsection{\textbf{\model changes its behavior across different embodiments}}
\label{sec:emb_aware}
% \vspace{-2mm}
%Our \model policy is trained across a large space of embodiments.
% The behavior of optimal navigation policy, $\pi^{*}_{\theta}(a_t \mid o_t^e,)$, should be strongly shaped by the agent’s body. For instance, 

Ideally, an optimal policy would modify its behavior depending on the embodiment. For instance, a thinner robot can navigate through narrow hallways or under furniture, and a wider agent may need to take more conservative paths. 
Our qualitative results show that \model exhibits embodiment-adaptive behavior. In Fig.~\ref{fig:bed_trajectory}-A,B, both Stretch RE-1 and Unitree A1 begin behind a bed. The low-profile quadruped moves directly under it, while Stretch RE-1 navigates around—demonstrating that \model \textit{implicitly} infers embodiment characteristics from visual input, without access to privileged body information. Visual input reveals cues like camera specs and, in some cases, the agent’s height. However, vision alone can be ambiguous, prompting the agent to rely on physical interactions—such as collisions—to refine its understanding. (Collision feedback may come from actual impacts or from sensors that anticipate collisions before they occur.) In Fig.~\ref{fig:bed_trajectory}-C, an agent with a low-mounted camera but tall body misjudges its own height, initially attempts to go under the bed, collides, and then reroutes like Stretch RE-1. This behavior is not present in the expert data but emerges during reinforcement learning through training across diverse embodiments. \textit{(Videos in supplementary)}

\input{tables/collision_comparison}
\subsection{\textbf{Collision analysis}}
\label{sec:ablations}

% \vspace{-2mm}
\model is trained with randomized body dimensions and is not explicitly provided with real embodiment information. Regardless, the learned policy remains as safe—\textit{and in some cases even safer}—than embodiment-specific policies (Tab.~\ref{tab:collision_comparison}). Collision-avoidance behavior can be further improved by incorporating a small collision penalty into the RL reward (Tab.~\ref{tab:collision_ablation}).

\noindent \textbf{\model learns to take conservative paths without explicit knowledge of its collider size.} Expert trajectories are generated using A*, finding minimum-cost paths where the cost is defined as the inverse of the Euclidean distance to the nearest obstacles. During reinforcement learning, collisions slow down the agent’s progress, leading to lower rewards through a step penalty. Since collisions cause no meaningful state changes and only waste time, the policy learns to avoid them to complete tasks more efficiently.
We evaluate \model against embodiment-specific baselines using the metric \textit{Safe Episodes}—the percentage of episodes completed without any collisions. As shown in Tab.~\ref{tab:collision_comparison}, \model achieves a higher percentage of Safe Episodes compared to the baselines. By training across a large number of embodiments and without access to exact body size information, \model learns to take more conservative actions through reinforcement learning, promoting safer navigation.

% When conservative paths are infeasible—e.g., in tight spaces or under low obstacles—the policy implicitly infers its collider size through collision feedback (Fig.~\ref{fig:bed_trajectory}). This behavior is not present in the expert data but is learned through reinforcement learning. 
% \ainaz{why?} \khz{A thought: could it because, during the RL training stage, collisions slow down the agent to accomplish tasks, resulting in lower rewards via step penalty? Collisions cause no state changes and completely waste time. Because of that, the policy learns to navigate safely (avoid collisions) to finish tasks?}

\noindent \textbf{Include collision penalty to take safer routes.} 
Adding a small collision penalty of $0.1$ to the reward function can further reduce collision rate (\textit{CR}) by $50\%$. The resulting policy is more conservative, regardless of embodiment size. To quantify these results, we created a custom benchmark similar to \textsc{Chores}-$\mathbb{S}$~\cite{ehsani2024spoc}, consisting of $2{,}000$ random embodiments across $2{,}000$ scenes. We evaluate $2$ different versions of our policy on this benchmark, comparing metrics such as \textit{Success Rate}, \textit{Success Weighted by Collision (SC)}, \textit{Collision Rate (CR)}, and \textit{Safe Episode}. As shown in Tab.~\ref{tab:collision_ablation}, adding the collision penalty reduces the collision rate (CR) ($7.77\%$ $\rightarrow$ $4.03\%$) as well as increases the percentage of trajectories without collisions ($46.90\%$ $\rightarrow$ $60.57\%$).

\input{tables/ablations_collision_penalty}

% \input{tables/main_results}

%% file: tables/zero_shot.tex
\begin{table*}[t]
\centering
\renewcommand{\arraystretch}{0.8}
\scriptsize
\addtolength{\tabcolsep}{-0.7em} % TODO added by Jordi - feel free to remove it
\vspace{-2mm}
\begin{tabular}{lcccccccc}
%\toprule
\multirow{2}{*}{Model} & \multirow{2}{*}{Loss} & \multirow{2}{*}{Train Embodiment} & \multicolumn{5}{c}{Benchmark Embodiment: Success (SEL)} &  \\
\cmidrule(r){4-8} %\cmidrule(r){4} \cmidrule(r){5}
& & & \multicolumn{1}{c}{\textbf{Stretch}} & \multicolumn{1}{c}{\textbf{Stretch} \tiny{(Nav Cam)}} & \multicolumn{1}{c}{\textbf{Stretch} \tiny{(Factory Config)}} & \multicolumn{1}{c}{\textbf{LoCoBot}} & \multicolumn{1}{c}{\textbf{Unitree A1}} & Average \\
\midrule
\multirow{1}{*}{SPOC~\cite{ehsani2024spoc}}  & \multirow{2}{*}{IL only} & \multirow{2}{*}{Stretch} & \textcolor{gray}{57.0 (38.1)*} & 37.9 (19.0) & 33.0 (19.3)& 16.2 (5.4) & 2.1 (1.6) & 29.2 (16.7) \\
\multirow{1}{*}{SPOC-$2.3$M}  &  & & \textcolor{gray}{60.0 (30.3)*} & 37.5 (17.9) & 46.0 (19.4) & 24.0 (7.9) & 10.0 (5.2)& 35.5 (16.1) \\
% %
% %
% %
\midrule
\multirow{3}{*}{\textsc{PoliFormer}~\cite{zeng2024poliformer}}  & \multirow{3}{*}{RL only} & Stretch & \textcolor{gray}{81.0 (58.1)*} & 65.0 (35.5) & 47.5 (25.6) & 27.5 (14.8) & 42.6 (25.1) & 52.7 (31.8) \\
 & & LoCoBot & 56.0 (32.9) & 56.5 (34.7) & 52.0 (27.7) & \textcolor{gray}{61.5 (44.7)*} & 50.5 (34.2) & 55.4 (34.9) \\
 & & Unitree A1 & 40.0 (25.2) & 39.0 (22.5) & 35.5 (20.9) & 30.0 (17.4) & \textcolor{gray}{55.3 (48.2)*} & 40.0 (26.8) \\
\midrule
\multirow{1}{*}{\textsc{FLaRe}~\cite{hu2024flare}}  & \multirow{1}{*}{IL + RL} & Stretch & \textcolor{gray}{82.0 (63.5)*} & 55.5 (37.9) & 38.0 (19.6) & 21.5 (10.9) & 27.0 (15.1) & 44.8 (29.4) \\
\midrule
\midrule
\modelzeroshot & \multirow{1}{*}{IL + RL} & \multirow{1}{*}{\model-Random} & \textbf{76.0 (55.9)} & \textbf{74.0 (52.5)} & \textbf{72.0 (52.7)}& \textbf{66.5 (45.3)} & \textbf{72.0 (58.6)} & \textbf{72.1 (53.0)} \\

%
%
%

%\bottomrule
\end{tabular}
\footnotesize

\caption{\footnotesize{\textbf{Zero-shot Results.} \model shows zero-shot generalization to four unseen embodiments. Unless otherwise specified, “Stretch” refers to the two-camera variant of the RE-1 platform, used in\cite{ehsani2024spoc}. All prior methods fail to generalize effectively to embodiments beyond those seen during training. \textcolor{gray}{Gray*} numbers indicate evaluation on the training embodiment; all others reflect zero-shot performance on unseen embodiments.}}
\label{tab:zero_shot}
\vspace{-2mm}
\end{table*}

%% file: tables/real_world.tex
% \begin{table}[t]
% \centering
% % \small
% \resizebox{\columnwidth}{!}{
% \begin{tabular}{lccccc}
% \multirow{2}{*}{Model} & \multirow{2}{*}{Train Embodiment} & \multicolumn{4}{c}{Eval Embodiment} \\
% \cmidrule(r){3-6}
% & & \multicolumn{1}{c}{\textbf{Stretch}} & \multicolumn{1}{c}{\textbf{Stretch (FC)}} & \multicolumn{1}{c}{\textbf{LoCoBot}} & \multicolumn{1}{c}{\textbf{Unitree Go1}} \\
% \midrule
% ProcTHOR~\cite{Deitke2022ProcTHORLE} & LoCoBot& - & - & \textcolor{gray}{26.7} & - \\
% \midrule
% Phone2Proc~\citep{Deitke2023Phone2Proc} & LoCoBot & - & - & \textcolor{gray}{66.7} & - \\
% \midrule
% SPOC~\cite{ehsani2023imitating} & Stretch& \textcolor{gray}{50.0} & - & - & - \\
% \midrule
% \multirow{3}{*}{\textsc{PoliFormer}~\cite{zeng2024poliformer}} & Stretch & \textcolor{gray}{83.3} & 33.3 & - & - \\
%  & LoCoBot & - & - & \textcolor{gray}{80.0} & - \\
%  & Unitree Go1 & - & - & - & \textcolor{gray}{41.7} \\
%  \midrule
% \textsc{FLaRE}~\cite{hu2024flare} & Stretch & \textcolor{gray}{\textbf{94.4}} & - & - & - \\
% \midrule
% \midrule
% \modelzeroshot & \model-Random & 83.3 & \textbf{72.2} & \textbf{80.0} & \textbf{80.0} \\
% % \modeltune & FPIN-Random & 83.3 & - & - & - \\
% % \bottomrule
% \end{tabular}
% }
% % \normalsize
% \caption{\textbf{Real-world Results}. \model transfers zero-shot to the real-world without any finetuning. \textcolor{gray}{Gray} numbers are evaluated on same embodiment as their training. \model achieves $78.9\%$ success rate on average across $4$ real-world robots.}
% \label{tab:real_world}
% \end{table}

\begin{table}[t]
\begin{minipage}{.5\linewidth}
  \centering
  
  \tiny
  \setlength{\tabcolsep}{0.8pt}
  \renewcommand{\arraystretch}{0.9}
    		\begin{tabular}{lccccc}
\multirow{2}{*}{Model} & \multirow{2}{*}{Train Embodiment} & \multicolumn{4}{c}{Eval Embodiment} \\
\cmidrule(r){3-6}
& & \multicolumn{1}{c}{\textbf{Stretch}} & \multicolumn{1}{c}{\textbf{Stretch (FC)}} & \multicolumn{1}{c}{\textbf{LoCoBot}} & \multicolumn{1}{c}{\textbf{Unitree Go1}} \\
\midrule
% ProcTHOR~\cite{Deitke2022ProcTHORLE} & LoCoBot& - & - & \textcolor{gray}{26.7} & - \\
% \midrule
% Phone2Proc~\citep{Deitke2023Phone2Proc} & LoCoBot & - & - & \textcolor{gray}{66.7} & - \\
% \midrule
SPOC~\cite{ehsani2023imitating} & Stretch& \textcolor{gray}{50.0} & - & - & - \\
\midrule
\multirow{3}{*}{\textsc{PoliFormer}~\cite{zeng2024poliformer}} & Stretch & \textcolor{gray}{83.3} & 33.3 & - & - \\
 & LoCoBot & - & - & \textcolor{gray}{80.0} & - \\
 & Unitree Go1 & - & - & - & \textcolor{gray}{41.7} \\
 \midrule
\textsc{FLaRE}~\cite{hu2024flare} & Stretch & \textcolor{gray}{\textbf{94.4}} & - & - & - \\
\midrule
\midrule
\modelzeroshot & \model-Random & 83.3 & \textbf{72.2} & \textbf{80.0} & \textbf{80.0} \\
% \modeltune & FPIN-Random & 83.3 & - & - & - \\
\midrule
\end{tabular}
\caption{\footnotesize{\textbf{Real-world Results}. \model transfers zero-shot to the real-world without any finetuning. \textcolor{gray}{Gray} numbers are evaluated on same training embodiment. }}
\label{tab:real_world}
\end{minipage}%
\hspace{0.5cm}
\begin{minipage}{.5\linewidth}
  \centering
  
  \tiny
 \setlength{\tabcolsep}{1pt}
  \renewcommand{\arraystretch}{0.7}
\begin{tabular}{lcccccccc}

\multirow{2}{*}{Model} & \multirow{2}{*}{Train Embodiment} & \multirow{2}{*}{Object} & \multicolumn{5}{c}{Human Participants} & \\
\cmidrule(r){4-8} %\cmidrule(r){4} \cmidrule(r){5}
& & & \multicolumn{1}{c}{\textbf{H1}}  & \multicolumn{1}{c}{\textbf{H2}} & \multicolumn{1}{c}{\textbf{H3}}& \multicolumn{1}{c}{\textbf{H4}}& \multicolumn{1}{c}{\textbf{H5}} & Average \\

\midrule
\multirow{3}{*}{\textsc{FLaRE}~\cite{hu2024flare}} & \multirow{3}{*}{Stretch RE-1} & \includegraphics[height=2mm]{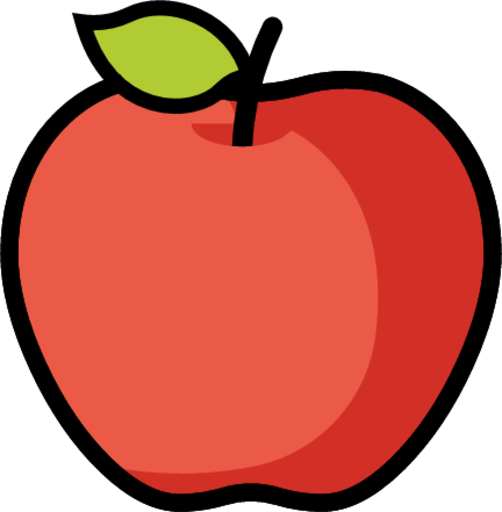} & \textcolor{OliveGreen}{\cmark} & \textcolor{BrickRed}{\xmark} & \textcolor{BrickRed}{\xmark} & \textcolor{BrickRed}{\xmark} & \textcolor{BrickRed}{\xmark} & \multirow{3}{*}{40.0\%} \\ 
&  & \includegraphics[height=2mm]{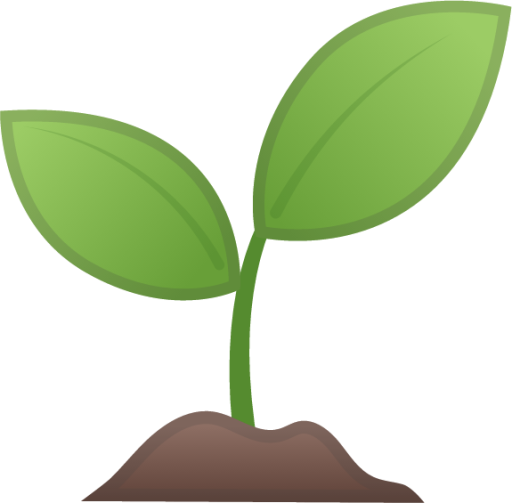} & \textcolor{BrickRed}{\xmark} & \textcolor{OliveGreen}{\cmark} & \textcolor{OliveGreen}{\cmark} & \textcolor{OliveGreen}{\cmark} & \textcolor{OliveGreen}{\cmark} & \\ 
&  & \includegraphics[height=2mm]{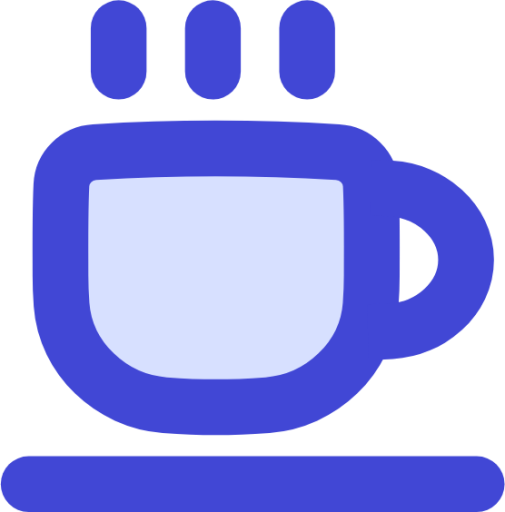} & \textcolor{BrickRed}{\xmark} & \textcolor{BrickRed}{\xmark} & \textcolor{BrickRed}{\xmark} & \textcolor{BrickRed}{\xmark} & \textcolor{OliveGreen}{\cmark}  & \\ 

\midrule
\multirow{3}{*}{\modelzeroshot} & \multirow{3}{*}{\model-Random} & \includegraphics[height=2mm]{figs/apple.png} & \textcolor{OliveGreen}{\cmark} & \textcolor{BrickRed}{\xmark} & \textcolor{BrickRed}{\xmark} & \textcolor{BrickRed}{\xmark} & \textcolor{OliveGreen}{\cmark} & \multirow{3}{*}{\textbf{73.3\%}} \\
& & \includegraphics[height=2mm]{figs/leaf.png} & \textcolor{OliveGreen}{\cmark} & \textcolor{OliveGreen}{\cmark} & \textcolor{OliveGreen}{\cmark} & \textcolor{OliveGreen}{\cmark} & \textcolor{OliveGreen}{\cmark} & \\
& & \includegraphics[height=2mm]{figs/mug.png} & \textcolor{BrickRed}{\xmark} & \textcolor{OliveGreen}{\cmark} & \textcolor{OliveGreen}{\cmark} & \textcolor{OliveGreen}{\cmark} & \textcolor{OliveGreen}{\cmark} & \\

% \bottomrule
\end{tabular}
\caption{\footnotesize{\textbf{Human Evaluation.} Five individuals navigate to 3 different objects ( \texttt{Apple}, \texttt{Houseplant},\texttt{Mug}) following the policy's output actions on their phones in a kitchen area (see Fig.~\ref{fig:human_eval}). }} 
\label{tab:humans}
\end{minipage}
\end{table}

%% file: figs/bed_trajectory.tex
\begin{figure*}
	\vspace{-3mm}
	%\hspace{4mm}
	\centering
	\includegraphics[width=0.9\textwidth]{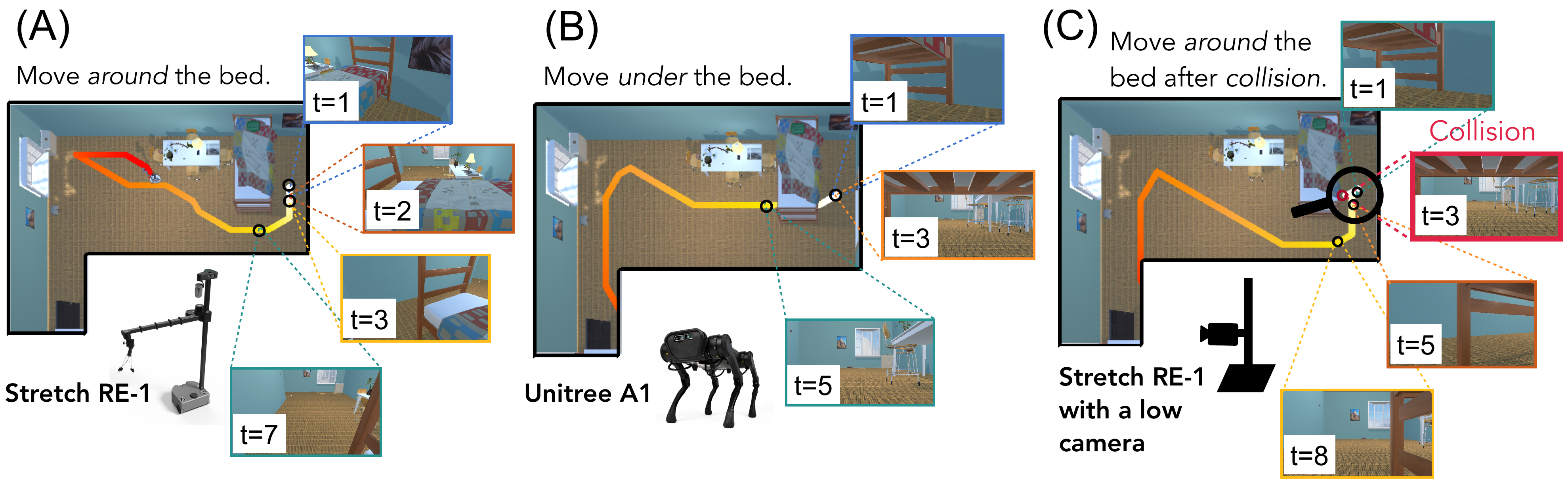}
\vspace{-2mm}
\caption{\footnotesize{
\model exhibits \textbf{embodiment-adaptive} behavior, adjusting its navigation strategy based on the robot’s physical configuration. The shorter quadruped robot (B) walks under the bed, while the taller Stretch-RE1 (A) navigates around it. In (C), an agent with the same height as Stretch-RE1 but a lower camera position initially attempts to go under the bed, mistakenly assuming it can fit. After a collision, it adapts and reroutes around the bed, similar to Stretch-RE1.
    }}
	\label{fig:bed_trajectory}
    \vspace{-6mm}
\end{figure*}

%% file: figs/finetune.tex
% \begin{figure}[t]
% 	\vspace{-1mm}
% 	%\hspace{4mm}
% 	\centering
% 	\includegraphics[width=1\columnwidth]{figs/finetune.pdf}
% 	\caption{
% \textbf{Embodiment-Specialized Adaptation.} \model, pretrained on randomized embodiments, shows efficient adaptation to robot-specialized policies with minimal fine-tuning. Baseline performance on LoCoBot and Unitree-A1 remains lower as they are fine-tuned on a different embodiment than the one used in pretraining. In contrast, \model policy achieves consistent performance across all 3 embodiments, highlighting its capability for robot-specialized adaptation.}
% 	\label{fig:finetune}
% \end{figure}

\begin{wrapfigure}{R}{0.58\textwidth}
\vspace{-7mm}
  \begin{center}
    \includegraphics[width=0.55\textwidth]{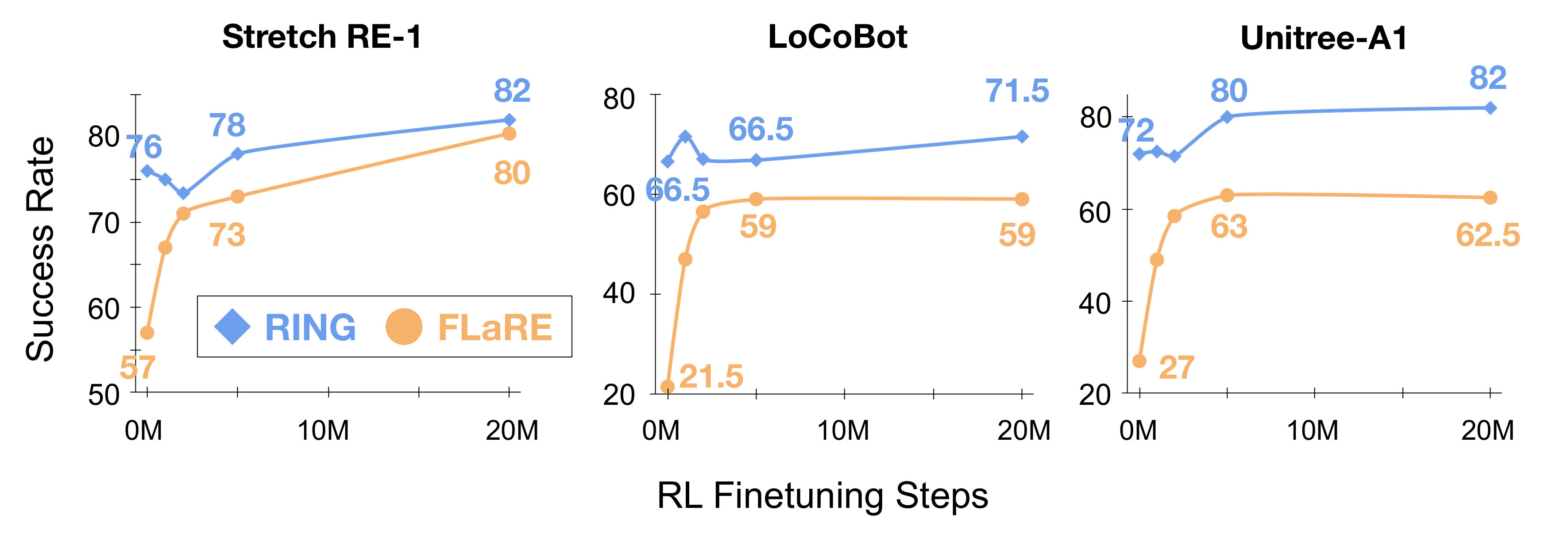}
  \end{center}
  \vspace{-4mm}
  \caption{\footnotesize{\textbf{Embodiment-Specialized Adaptation.} \model, pretrained on randomized embodiments, adapts efficiently to individual robots with minimal fine-tuning.  }}

	\label{fig:finetune}
  \vspace{-1mm}
\end{wrapfigure}

%% file: tables/collision_comparison.tex
\begin{table}[t]
\begin{minipage}{.5\linewidth}
  \centering
  
  \vspace{-2mm}
  \tiny
  \setlength{\tabcolsep}{1pt}
  \renewcommand{\arraystretch}{0.8}
\begin{tabular}{lcccc}
%\toprule
\multirow{2}{*}{Model} & \multirow{2}{*}{Train Embodiment} &  \multicolumn{3}{c}{Safe Episode $\uparrow$} \\
\cmidrule(r){3-5} %\cmidrule(r){4} \cmidrule(r){5}
& & \multicolumn{1}{c}{\textbf{Stretch}}  & \multicolumn{1}{c}{\textbf{LoCoBot}} & \multicolumn{1}{c}{\textbf{Unitree A1}} \\
\midrule
\multirow{3}{*}{\textsc{PoliFormer}~\cite{zeng2024poliformer}} & \multirow{1}{*}{Stretch} & 45.0 & - & - \\
 & \multirow{1}{*}{LoCoBot} & - & 42.0 & - \\
 & \multirow{1}{*}{Unitree A1} & - & - & \textbf{49.25} \\

 \midrule

\multirow{1}{*}{\textsc{FLaRe}~\cite{hu2024flare}}& \multirow{1}{*}{Stretch} & 62.0 & - & - \\
 \midrule
 \midrule

\multirow{1}{*}{\modelzeroshot} & \multirow{1}{*}{\model-Random} & \textbf{67.0} & \textbf{64.5} & 48.5 \\

\midrule

\end{tabular}
\caption{\footnotesize{ \model has more \textbf{Safe Episodes} (with no collisions) compared to embodiment-specific baselines.}}
\label{tab:collision_comparison}
\end{minipage}%
\vspace{-2mm}
\hspace{0.1cm}
\begin{minipage}{.45\linewidth}
  \centering
  
  \vspace{-2mm}
  \scriptsize
  \setlength{\tabcolsep}{0.8pt}
  \renewcommand{\arraystretch}{1.}
\begin{tabular}{lcccccc}
%\toprule
\multirow{2}{*}{Model} & \multirow{2}{*}{Collision Penalty} & \multicolumn{5}{c}{Metrics}  \\
\cmidrule(r){3-7}
& & Success $\uparrow$ & SEL $\uparrow$ & SC $\uparrow$ & CR $\downarrow$ & Safe Episode $\uparrow$ \\
\midrule
\multirow{2}{*}{\textit{\model}} & \textcolor{BrickRed}{\xmark} & \textbf{67.62} & 56.24 & 42.53 & 7.77 & 46.90 \\
& \textcolor{OliveGreen}{\cmark} & 66.33 & \textbf{56.87} & \textbf{49.05} & \textbf{4.03} & \textbf{60.57} \\
\midrule
\end{tabular}
\caption{\footnotesize{\textbf{Collision Penalty.} Adding a small collision penalty (0.1) to the reward function results in 50\% less collision, forcing the policy to be even more conservative.}} 
\label{tab:collision_ablation}
\end{minipage}
\vspace{-5mm}
\end{table}

%% file: tables/ablations_collision_penalty.tex
% \setlength{\tabcolsep}{3pt}
% \begin{table}[t]
% \centering
% \renewcommand{\arraystretch}{1.0}
% \scriptsize
% \begin{tabular}{lcccccc}
% %\toprule
% \multirow{2}{*}{Model} & \multirow{2}{*}{Collision Penalty} & \multicolumn{5}{c}{Metrics}  \\
% \cmidrule(r){3-7}
% & & Success $\uparrow$ & SEL $\uparrow$ & SC $\uparrow$ & CR $\downarrow$ & Safe Episode $\uparrow$ \\
% \midrule
% \multirow{2}{*}{\textit{\model}} & \textcolor{BrickRed}{\xmark} & \textbf{67.62} & 56.24 & 42.53 & 7.77 & 46.90 \\
% & \textcolor{OliveGreen}{\cmark} & 66.33 & \textbf{56.87} & \textbf{49.05} & \textbf{4.03} & \textbf{60.57} \\
% \midrule
% \end{tabular}
% \normalsize
% \caption{\textbf{Collision Penalty.} Adding a small collision penalty (0.1) to the reward function results in 50\% less collision, forcing the policy to take more conservative paths.}
% \label{tab:collision_ablation}
% \vspace{-1.5em}
% \end{table}

%% file: sec/5_conclusion.tex
\section{Conclusion}
\vspace{-3mm}

In this paper, we introduce \model (\fullname), an embodiment-agnostic policy trained entirely in simulation on 1 million diverse, randomly initialized embodiments. \model demonstrates strong zero-shot generalization to unseen embodiments, maintaining consistent performance across a wide range of robots. It achieves state-of-the-art results on novel embodiments—sometimes even outperforming embodiment-specific policies—and transfers directly to the real world despite never seeing real robot embodiments during training. Finally, \model dynamically adapts its behavior based on its embodiment and interactions with the environment.

% In this paper, we introduce \model (\fullname), an embodiment-agnostic policy, trained solely in simulation with diverse randomly initialized embodiments at scale (1M embodiments). \model displays zero-shot generalization capability to various unseen embodiments,  maintaining consistent performance across all. Our experimental results demonstrate that \model achieves state-of-the-art results on novel embodiments, including in some cases improving over embodiment-specific policies. It can be directly deployed to the real-world despite being solely trained in simulation. Finally, \model is able to dynamically adjust its behavior based on its embodiment and interactions with the environment.

%% file: sec/6_limitations.tex
\section{Limitations}
\label{app:parameters_conditioning}

Although \model has the advantage of being deployable on a wide range of embodiments without any privileged information about its current body, when available it may be beneficial to have a policy explicitly conditioned on the current embodiment specification. This might lead to improved performance and more desirable behaviors, such as increased efficiency and collision avoidance.

We train \modelembodimentconditioned by explicitly providing the embodiment information to the policy. The embodiment parameters are represented as a configuration vector $\mathbf{c}_e \in \mathbb{R}^{19}$, with each dimension corresponding to a specific embodiment parameter listed in Table~\ref{tab:body_params}. This information is passed as an additional token to the Transformer State Encoder. We use a simple MLP to project $\mathbf{c}_e$ to the desired feature dimension ${e \in \mathbb{R}^{1 \times 512}}$ before passing it to the encoder. Tab.~\ref{tab:ablations_agent_params} evaluates the 2 versions of the policy on our custom benchmark consisting of $2{,}000$ random embodiments across $2{,}000$ scenes, comparing metrics such as \textit{Success Rate}, \textit{Success Weighted by Collision (SC)}, \textit{Collision Rate (CR)}, and \textit{Safe Episode (percentage of episodes without any collisions)}.

The results do not show a clear benefit to conditioning the policy on embodiment information. This could be due to several reasons. 
% It is possible that the environment is not sufficiently complex for the policy to require explicit information, and the policy can instead implicitly infer the embodiment configuration from visual observations and collisions. 
It is possible that most relevant information about environment hazards and agent motion can be already inferred from visual observations. It is also possible that a significant fraction portion of collisions (both with an without embodiment specification provided) occur with objects that never enter the agent's visual field, in which case extra information about its own embodiment would not help. Alternatively, a more effective method for conditioning the policy on the parameters may exist. Future work should explore this 
% by using more complex environments 
with additional examination of agent-environment collision and designing improved policy architectures to better integrate embodiment parameters, ultimately training a more efficient and robust policy that explicitly incorporates embodiment information.

 \input{tables/ablations_agent_param}

%% file: tables/ablations_agent_param.tex
\setlength{\tabcolsep}{2pt}
\begin{table}[H]
\centering
\renewcommand{\arraystretch}{1.0}
\scriptsize
\begin{tabular}{lcccccc}
%\toprule
\multirow{2}{*}{Model} & \multicolumn{1}{c}{Ablations} & \multirow{2}{*}{Success $\uparrow$} & \multirow{2}{*}{SEL $\uparrow$} & \multirow{2}{*}{SC $\uparrow$} & \multirow{2}{*}{CR $\downarrow$}  & \multirow{2}{*}{Safe Episode $\uparrow$} \\
\cmidrule(r){2-2} 
&  \multicolumn{1}{c}{Embodiment Parameters} & & & & &  \\
\midrule
\multirow{1}{*}{\textit{\model}}  & \multirow{1}{*}{\textcolor{BrickRed}{\xmark}} &  67.62 &  56.24 & 42.53 & \textbf{7.77} & \textbf{46.90} \\
\multirow{1}{*}{\textit{\modelembodimentconditioned}} & \multirow{1}{*}{\textcolor{OliveGreen}{\cmark}}  &  \textbf{69.44} &  \textbf{57.42} & \textbf{44.69} & 8.0 & 46.54 \\

\bottomrule
\end{tabular}
\normalsize
\vspace{2mm}
\caption{\footnotesize{\textbf{Conditioning \model on embodiment parameters}. We explicitly provide the embodiment parameters to the policy (\modelembodimentconditioned) and compare with \model without any information about the embodiment. Both policies are evaluated on a custom benchmark consisting of 2000 random embodiments in 2000 scenes.}}
\label{tab:ablations_agent_params}
\vspace{-1em}
\end{table}

%% file: sec/X_suppl.tex
\clearpage
% \setcounter{page}{1}
% \maketitlesupplementary

% \twocolumn[
% \begin{center}
%     {\LARGE {\textbf{Appendix}} \\ \vspace{4mm} The One RING\includegraphics[height=4mm]{figs/logo_ring_v2.png}: a Robotic Indoor Navigation Generalist}
%     \vspace{0.2cm} % Adjust spacing below the title
% \end{center}
% \vspace{1cm} % Add space before starting the two-column content
% ]
% \section*{Appendices for \emph{The One RING \includegraphics[height=5mm]{figs/logo_ring_v2.png}: a Robotic Indoor Navigation Generalist}}
\section*{Appendices for \emph{The One RING \includegraphics[height=5mm]{figs/logo_ring_v2.png}: a Robotic Indoor Navigation Generalist}}

The following items are provided in the Appendix:
\textit{
\begin{itemize}
    \item Details about the real-world evaluation robot platforms and human evaluation setup (App.~\ref{app:real_world_details}),
    \item Data collection for randomized embodiments using expert planners in simulation (App.~\ref{app:experts}),
    \item Full experimental setup (App.~\ref{app:experiment_details}),
    \item Model architecture details (App.~\ref{app:model_arch_details}),
    \item We study the impact of using a more powerful visual encoder (App.~\ref{app:visual_encoder}),
    \item A visualization of the random embodiments in our training set, along with the 5 nearest neighbors to each of the real robots (Stretch-RE1, LoCoBot, Unitree A1) (App.~\ref{app:tsne_nn}),
    \item Out-of-distribution generalization for different embodiment parameters (App.~\ref{app:ood_generalization})
\end{itemize}
}

% Please find our project website (see the \href{https://poliformer.allen.ai}{poliformer.allen.ai}) that contains 
% \noindent On our website (\href{https://one-ring-policy.allen.ai/}{\textcolor{darkpink}{one-ring-policy.allen.ai}}), we have
\noindent On our website (\href{https://one-ring-policy.allen.ai/}{\textcolor{darkpink}{one-ring-policy.allen.ai}}), we have
\textit{
\begin{itemize}
\item Real-world qualitative videos of evaluating \model zero-shot on five different robot platforms, including Stretch RE-1 with our camera setup, Stretch RE-1 with factory camera configuration, LoCoBot, Unitree GO1, and RB-Y1 wheeled humanoid,
\item Qualitative videos for human evaluation, using \model as navigation assistant,
\item Videos showing our dataset of trajectories collected from random embodiments in simulation.
\end{itemize}
}

\input{figs/real_robot_layouts}
\section{Real Robot Platforms and Human Evaluation Setup.}
\label{app:real_world_details}

\subsection{Stretch RE-1, LoCoBot, Unitree GO1, RB-Y1}
\label{app:real_robot_details}
We use Stretch RE-1, LoCoBot, Unitree GO1, and RB-Y1 wheeled humanoid as our robot platforms for real-world evaluations, shown in Fig.~\ref{fig:robot_platforms}. For Stretch RE-1, we evaluate two different sensor configurations: the factory configuration and the configuration suggested by SPOC~\cite{ehsani2024spoc}. For RB-Y1, we include both standing and seated configurations and use an iPhone 16 Pro camera mounted on the robot to stream visual observations. The main differences among these platforms are summarized in Tab.~\ref{tab:robot_spec}. For robot movements, we either implement a Kalman filter or wrap around provided robot APIs to realize low-level controllers for a discrete action space \{\texttt{MoveBase(±20cm)}, \texttt{RotateBase(±$6^\circ$, ±$30^\circ$)}, \texttt{Done}\} across all platforms. It is important to note that during the training stage, we do not use any embodiment configurations from these robots to generate imitation learning data or to initialize RL fine-tuning embodiments.
\input{figs/human_eval_app}

\input{tables/robot_platforms}
\input{tables/human_spec}

\subsection{Human Evaluations}
\label{app:human_eval_details}
\noindent\textbf{Human participants}. 
We asked five human participants to use \model as a navigation assistant and evaluated its performance across a diverse range of human embodiments. These embodiment variations arose from differences in camera-holding posture, participant height, step size, and rotation angle. Tab.~\ref{tab:human_spec} summarizes these variations across participants. As a result, each participant contributed a distinct set of evaluation embodiments and sensor configurations.

\noindent\textbf{Human Evaluation Details.} We developed a simple iOS app (Fig.~\ref{fig:human_eval_app}) that allows human participants to input a text prompt (e.g., “Find a mug”), capture an image using the iPhone’s back camera, and send both the prompt and image to a remote server. The server processes this input using our \model policy, predicts action probabilities, samples an action, and returns it to the app for display. The action space available in the app mirrors that of our real-world robot.

Participants follow the suggested action at their own pace and chosen rotation degree, as specified in Tab.~\ref{tab:human_spec}. After each step, they tap the \texttt{Predict} button to repeat the process: capturing a new image and sending it, along with the original prompt, to the server. This continues until either the \texttt{Done} action is returned or 100 steps have been executed. An episode is considered successful if the target object is visible in the final image, within 1 meter, when \model issues the \texttt{Done} action.

Fig.~\ref{fig:human_eval} shows the layout of the evaluation scene and two sample trajectories, including two locations for a \texttt{Mug}, three for a \texttt{Houseplant}, and one for an \texttt{Apple}. Participants always begin in the bottom-left corner of the scene (results shown in Tab.~\ref{tab:humans}).

\input{figs/robot_platforms}

\input{figs/rainbow_traj}
\input{figs/human_eval}

\section{Data Generation with Expert Planners}
\label{app:experts}

Expert planners introduced by ~\cite{ehsani2024spoc} are not efficient and robust for random embodiments. As a result, we made major improvements to the planners to allow for better trajectories.

The major factor in this improvement is to consider \emph{safety} of the policy (defined as the avoidance of approaching any obstacles along the way.) We use A* \cite{hart1968astar,hagberg2008networkx} to generate safe navigation trajectories for training as follows:
% \textbf{1)} Extract reachable locations in a scene on a finely-spaced grid, such that the agent's collider does not intersect with any object's collider along a Manhattan path from the initial agent's pose to each reachable location. The adaptation to different embodiments is thus given by the interaction of the corresponding colliders and the environment.
\textbf{1)} Extract reachable locations in a scene on a finely spaced grid, ensuring that the agent’s collider does not intersect with any object’s collider. Thus, different embodiments yield different reachable locations according to their collider.
%\textbf{2)} Compute a clipped Euclidean distance transform to the nearest obstacle, and define the cost of visiting each location as the inverse of the third power of the corresponding distance.
\textbf{2)} Compute a clipped Euclidean distance to the nearest obstacle. Then, for each location, set the cost of visiting it as the inverse of the third power of the distance.
% \textbf{3)} Construct a graph with 4-connectivity, where each vertex is a reachable location and the edge costs are determined as the maxima of the costs of visiting each of the two connected locations.
\textbf{3)} Construct a grid-like graph where each reachable location is a node connected to its immediate neighbors. For each connection, assign a cost equal to the maximum cost of visiting either of the two connected nodes.
\textbf{4)} Extract a minimum-cost path connecting the reachable positions in the graph nearest to the source and to the target via A*.
%\textbf{5)} Extract waypoints by skipping locations (vertices) in the A* path as long as the cost of visiting the discrete locations along the resulting path segment does not increase the path cost from the latest waypoint.
\textbf{5)} Extract waypoints by skipping over points in the A* path as long as skipping them doesn’t increase the total path cost from the latest waypoint.
\textbf{6)} The expert linearly interpolates between waypoints up to the precision reachable by the action space to generate each trajectory.

\section{Additional Benchmark/Experiment Details}
\label{app:experiment_details}
\noindent\textbf{Action Space.} Following on prior work with AI2-THOR, we discretize the action space for all agents in our training: \{\texttt{MoveAhead, MoveBack, RotateRight, RotateLeft, RotateRightSmall, RotateLeftSmall, Done}\}. Here, \texttt{MoveAhead} advances the robot by 0.2 meters, \texttt{MoveBack} moves the robot backward by 0.2 meters, \texttt{RotateRight/RotateRightSmall} rotates it clockwise by $30^\circ$ /  $6^\circ$ around the yaw axis, and \texttt{RotateLeft/RotateLeftSmall} rotates it counterclockwise by $30^\circ$ /  $6^\circ$ around the yaw axis, and \texttt{Done} indicates the agent has located the target, ending the episode.
We evaluate \model zero-shot on all robots (Stretch-RE, LoCoBot, Unitree Go1) with the same action space using their low-level controllers.
When finetuning for embodiment-specialized policies, we finetune for a slightly different action space for LoCoBot: \{\texttt{MoveAhead, MoveBack, RotateRight, RotateLeft, LookUp, LookDown, Done}\}. \texttt{LookUp} tilts the camera up by $30^\circ$ around the roll axis and \texttt{LookDown} tilts the camera down by $30^\circ$ around the roll axis. All baselines are trained and evaluated with the same action space for fair comparison.

\noindent\textbf{Success Criteria.} We follow the definition of Object Goal Navigation from \cite{Batra2020ObjectNav}, where an agent must explore its environment to locate and navigate to a specified object within a maximum of $n$ steps. To indicate it has found the target, the agent must execute the \texttt{Done} action. Success is determined by the environment based on whether the agent is within a distance $d$ of the target and if the target is visible in its view. If the agent exceeds $n$ steps without executing the \texttt{Done} action, the episode is considered a failure. For simulation benchmarks, we follow CHORES-S~\cite{ehsani2024spoc} with $n=600$ and $d=2$. For real-world evaluations, we use $n=300$ and $d=1$.

\noindent\textbf{Success weighted by collision (SC).} Collision is one of the main challenges for a unified policy operating across diverse embodiments in visual navigation tasks. Previous works measure the collision rate ($\frac{\# collisions}{\# steps}$) to understand how often a policy collides with objects in a scene. However, this does not reflect the effectiveness of the policy at the task level. For example, in a successful episode, a single collision and multiple collisions should have different impacts on the performance measurement. As a results, inspired from Success Weighted by Episode Length (SEL), we propose Success Weighted by Collision (SC),
\vspace{-3mm}
\begin{equation}
    SC = \frac{1}{N}\sum_{i=1}^{N}S_{i}\frac{1}{1 + c_i},
\end{equation}
where $S_i$ is a binary indicator of success for episode $i$, $c_i$ is the number of collisions in episode $i$, and $N$ is the number of evaluation episodes. In this metric, the policy is penalized most heavily for a single collision, with the penalization decreasing for each additional collision, as the penalty diminishes inversely with the number of collisions. Intuitively, $>0$ collisions are much worse than $0$, as a real robot may suffer damage from one bad collision, but the difference between $10$ and $11$ collisions is a more marginal difference.

\noindent\textbf{Hyparameters.}
We list the hyperparameters used in training and the architecture in Table~\ref{tab:hyperparams}.
\input{tables/hyperparams}

\input{figs/model_details}
\subsection{Real-World Benchmarks}
We evaluate 3 of the robots (Stretch-RE1, LoCoBot, Unitree Go1) in a multi-room apartment shown in Fig. \ref{fig:real_layout}. Based on the embodiment, the benchmark has different starting locations and objects. Among our target object categories, \texttt{Apple} can be found in the Living room and Kitchen, \texttt{Bed} can only be found in the Bedroom, \texttt{Sofa} and \texttt{Television} can only be found in the Living room, \texttt{Vase} can be found in the Livingroom, Corridor, Office, and Kitchen, \texttt{Chair} can be found in the Office and Kitchen, \texttt{HousePlant} can be found in the Living room, Office, and Kitchen.

We also deploy the policy on the RB-Y1 wheeled humanoid in an unstructured kitchen environment—similar to the human evaluation setting.

\begin{itemize}
    \item \textbf{LoCoBot}: Following Phone2Proc~\citep{Deitke2023Phone2Proc}, use the same five target object categories, including \texttt{Apple}, \texttt{Bed}, \texttt{Sofa}, \texttt{Television}, and \texttt{Vase}, and the three starting poses shown in \ref{fig:real_layout}.
    \item \textbf{Stretch RE-1}: We follow SPOC~\citep{ehsani2023imitating} to use the same six target object categories, including \texttt{Apple}, \texttt{Bed}, \texttt{Chair}, \texttt{HousePlant}, \texttt{Sofa}, and \texttt{Vase}, and the three starting poses, shown in Fig.~\ref{fig:real_layout}. We consider 2 different camera configurations for Stretch: 1) off-the-shelf camera equipped on the Stretch RE-1 (D435 with a vertical field of view of $69^{\circ}$ and resolution of $720 \times 1280$), 2) following \cite{ehsani2024spoc}, we use 2 Intel RealSense 455 fixed cameras, with a vertical field of view of $59^{\circ}$ and resolution of $1280 \times 720$. The cameras are mounted facing forward but pointing downward, with the horizon at an angle of $27^{\circ}$. 
    \item \textbf{Unitree Go1}: We create a new evaluation set for Unitree Go1 with 3 starting poses (Fig.~\ref{fig:real_layout}) and 4 objects (\texttt{toilet, sofa, TV, trashcan}) positioned to accommodate the robot's lower height, ensuring that the objects can be visible from its lower viewpoint. 
    \item \textbf{RB-Y1}: The wheeled humanoid navigates to various target objects in a real kitchen area, including a \texttt{mug}, \texttt{apple}, \texttt{houseplant}, and \texttt{trashcan} (2 example trajectories shown in Fig.~\ref{fig:rainbow_traj}).
\end{itemize}

% See \textit{Details about Different Benchmarks} in page 22 at \url{https://openreview.net/pdf?id=KdVLK0Wo5z#page=22.78}.

\section{Model Architecture Details}
\label{app:model_arch_details}

We will now detail \model's architecture (see Fig.~\ref{fig:model_details}), which is inspired by previous works \textsc{PoliFormer}~\cite{zeng2024poliformer} and \textsc{FLaRe}~\cite{hu2024flare}. 

% We train two versions of our policy: \model (without explicit embodiment information) and \modelembodimentconditioned, where the policy $\pi_\theta(a_t \mid o_t, \mathbf{c}_e)$ is explicitly conditioned on the embodiment parameters. In order to incorporate these parameters, we pass the $\mathbf{c}_e$ vector as embodiment tokens to the transformer state encoder along with the visual representations and the goal features.

% We make a few changes to the architecture: we replace the DINOv2~\cite{oquab2023dinov2} visual backbone with the visual encoder introduced in Molmo~\cite{deitke2024molmo} trained on the dense-captioning data PixMo-Cap~\cite{deitke2024molmo}. Changing to this visual encoder has dramatic effect on improving the performance.

\noindent\textbf{Visual encoder}. 
\label{sec:visual-encoder} We use the Vision Transformer from the pretrained \textsc{SIGLIP-ViT-B/16} as our visual encoder. Since the RGB images vary in dimensions across different embodiments, we include an additional preprocessing step before feeding them into the encoder. Specifically, we pad each RGB image to a square and then resize it to $256\times256$. In addition, we mask the image from the $2^{nd}$ camera with zeros for the embodiments with only one camera. The visual backbone takes the RGB observation $i \in \mathbb{R}^{256\times 256\times 3}$ as input and produces a patch-wise representation $r \in \mathbb{R}^{\frac{256}{16}\times \frac{256}{16}\times h}$, where $h=768$ is the hidden dimension of the visual representation. We reshape the visual representation into a $\ell\times h$ matrix, $\ell=256\cdot 256/16\cdot 16$, and project the representation to produce $v \in \mathbb{R}^{\ell\times d}$, where $d=512$ is the input dimension to the transformer state encoder. Note that since we have two RGB images from two cameras, we produce two visual representations $v^{1,2}$ at the end of this module. The vision encoder remains frozen through training.

% Molmo~\cite{deitke2024molmo} visual encoder which is the OpenAI’s ViT-L/14 336px CLIP model~\cite{radford2021learning} trained on PixMo-Cap~\cite{deitke2024molmo} dense-captioning dataset. The dense-captioning representation can identify a large number of object categories, their attributes, their spatial relationships, etc. providing valuable information for solving the task. 

\noindent\textbf{Goal encoder.}
\label{sec:goal-encoder} We follow the Text Transformer from the pretrained \textsc{SIGLIP-ViT-B/16} to encode the given natural language instruction into goal embedding $t \in \mathbb{R}^{64\times h}$, where $h=768$ is the hidden dimension and this Text Transformer returns $64$ tokens after padding. Before passing the goal embedding to the transformer state encoder, we always project the embedding to the desired dimension $d=512$, resulting in $g \in \mathbb{R}^{64\times 512}$.

% \noindent\textbf{Embodiment Parameter Encoder (Optional).}
% \label{sec:embodiment-encoder} 
% To train the \modelembodimentconditioned policy, we represent the embodiment parameters $\mathbf{c}_e \in \mathbb{R}^{19}$ as 19 individual tokens each corresponding to a specific embodiment parameter listed in Table.~\ref{tab:body_params}. A simple MLP is used to project each token to the desired feature dimension $\{e_i \in \mathbb{R}^{1\times d}\}_{i=1}^{19}$.

\noindent \textbf{Transformer State Encoder.}
This module summarizes the state at each timestep as a vector $s\in \mathbb{R}^{d}$. The input to this encoder includes two visual representations $v^{1,2}$, the goal feature $g$, and an embedding $f$ of a \texttt{STATE} token. These features are concatenated and fed to the non-causal transformer encoder. The output corresponding to the \texttt{STATE} token is the state feature vector $s\in \mathbb{R}^{d}$ which summarizes the state at each timestep. This feature vector is a goal-conditioned visual state representation.

\noindent \textbf{Causal transformer decoder.}\label{sec:causal-transformer}
We use a causal transformer decoder to perform explicit memory modeling over time. %\jordi{should we be careful with the explicit memory naming? I'm not sure if that could be interpreted as some sort of dedicated memory structure designed for navigation. If it seems actually clear feel free to ignore}
This can enable both long-horizon (e.g., exhaustive exploration with backtracking) and short-horizon (e.g., navigating around an object) planning.
Concretely, the causal transformer decoder constructs its state belief $b^{t}$ using the sequence of state features $\textbf{s} = \{s^j|_{j=0}^{j=t}\}$ within the same trajectories. To avoid recomputing the attention on the previous state features, we follow PoliFormer~\cite{zeng2024poliformer} to use KV-Cache to store the past \textbf{K}ey and \textbf{V}alue into two cache matrices in each attention layer. Therefore, we only perform feedforward computation for the most current state feature $s^{t}$.

\noindent \textbf{Linear actor-critic head.}
With the latest state belief $b^{t}$, we simply use a linear actor-critic head to project it to predict action logits over the action space. For RL-finetuning, the linear actor-critic head also predicts a value estimate about the current state.

\input{tables/nearest_neighbors_stretch}
\input{tables/nearest_neighbors_unitree}

\section{A More Powerful Pretrained Visual Encoder.}
\label{app:visual_encoder}
The default vision encoder used in our policies is the pretrained \textsc{SIGLIP-ViT-B/16}. In this section, we examine the impact of using a more powerful visual encoder on \model's performance. We train \modellarge using OpenAI’s \textsc{ViT-L/14 336px} CLIP model~\cite{radford2021learning}. Table~\ref{tab:vision_encoder_ablation} compares the results, showing that a stronger visual encoder significantly improves zero-shot performance across all four embodiments (approximately $9\%$ improvement on average). A larger visual encoder is particularly beneficial in our policy, as the visual observations are highly varied due to randomized camera parameters. To ensure fair comparison with the baselines and because \textsc{ViT-L/14} is more computationally demanding, we chose to use the \textsc{ViT-B/16} encoder for our main experiments. We will release the training code for the community for those interested in training with the larger visual encoder.
%\vspace{0.25cm}\\
\vspace{-2mm}
\input{tables/ablations_vision_encoder}

\input{figs/tsne_nn}
\section{Nearest Neighbor Embodiments to Real Robots in our Training Data}
\label{app:tsne_nn}

\input{figs/random_embodiments}

Fig.~\ref{fig:tsne_nn} presents a t-SNE visualization of the embodiment parameters $\mathbf{c}_e \in \mathbb{R}^{19}$ for 50k samples from the random embodiments in our training set (examples showin in Fig.~\ref{fig:random_embodiments}). We also show the corresponding parameters for Stretch, LoCoBot, and Unitree A1 for visualization purposes. Our random embodiments range widely over the space of possible embodiments, with many closely approximating each of the three real robots. Tables~\ref{tab:nn_stretch}, \ref{tab:nn_locobot}, and \ref{tab:nn_unitree} list the five nearest neighbors to each robot in the compressed t-SNE space and their corresponding embodiment parameters. Although the nearest neighbors do not exactly match each robot's embodiment, they are sufficiently similar across different parameters. This extensive coverage of the embodiment space and proximity to real-world embodiments ensure consistent zero-shot generalization to all three robots.

% It is worth noting that some parameters of the Unitree A1 fall outside the distribution of the training data. For example, we sample the collider size within the range $[0.2m, 0.5m]$ for both the x and z axes, whereas the Unitree A1 has a length of $0.64m$. This demonstrates that \model has the potential for out-of-distribution generalization.

\section{Generalization to out-of-distribution embodiment parameters}
\label{app:ood_generalization}
\input{figs/ood_ranges}
\input{tables/ood_generalization}
The random embodiments in our training set span a wide range of possible configurations, with many closely approximating each of the three real robots. Although the training data covers the full range of each embodiment parameter individually, the specific combination of parameters corresponding to each real robot is not explicitly included. This is demonstrated by the nearest-neighbor embodiments shown in Appendix~\ref{app:tsne_nn}. In this section, we examine the extent to which the policy generalizes to out-of-distribution values of individual embodiment parameters.

We focus on four specific parameters: \textit{camera height, camera field of view (FOV), camera pitch}, and \textit{collider size}. For each parameter, we define a narrower range that excludes the values corresponding to the real robots. From the training data, we filter the random embodiments to select 50k samples within each of these specified ranges. For comparison, we also train a version of the policy using 50k unfiltered embodiments that span the full range of each parameter. The selected training ranges for each parameter are illustrated in Fig.~\ref{fig:ood_ranges}.

We then perform zero-shot evaluations of the policies trained on each selected range using four robots whose parameters lie outside the training ranges. The success rate and collision rate are summarized in Table~\ref{tab:ood_results}. The results indicate that policies trained on narrower ranges still generalize to out-of-distribution parameters, achieving only a slightly lower success rate. However, evaluation on unseen embodiment parameters leads to a significantly higher collision rate, particularly for the policy trained with a narrower range of collider sizes. This suggests that the agent may rely more on physical contact with the environment to infer its embodiment configurations. Comparing Table~\ref{tab:ood_results} and Table~\ref{tab:zero_shot}, the average success rate drops by 13\%, emphasizing that the number of random embodiments used during training is crucial to develop an embodiment-agnostic policy capable of effectively handling a wide range of embodiments.

%% file: figs/real_robot_layouts.tex
% \begin{figure}
% 	%\hspace{4mm}
% 	\centering
% 	\includegraphics[width=0.9\columnwidth]{figs/real_world_layouts.pdf}
% 	\caption{
% \textbf{Real Evaluation Environment}. Our real-world evaluations are performed in a multi-room apartment with a long corridor, shown here with the three starting locations for three different robots' evaluations.}
% 	\label{fig:real_layout}
% \end{figure}

\begin{wrapfigure}{R}{0.45\textwidth}
\vspace{-7mm}
  \begin{center}
    \includegraphics[width=0.45\textwidth]{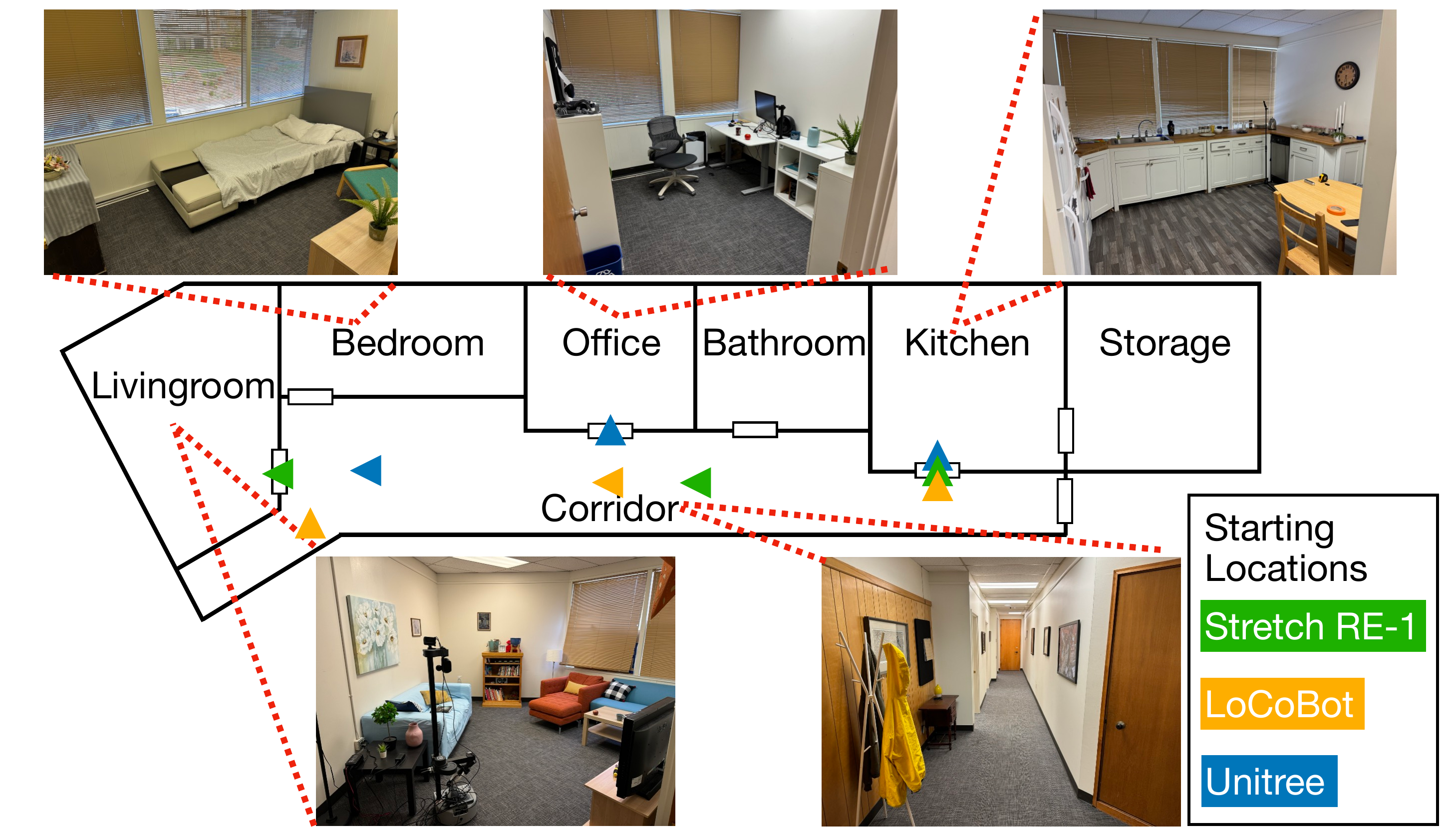}
  \end{center}
  \vspace{-4mm}
  \caption{\footnotesize{\textbf{Real Evaluation Environment}. Our real-world evaluations are performed in a multi-room apartment with a long corridor, shown here with the three starting locations for three different robots' evaluations.}}
	\label{fig:real_layout}
  % \vspace{-4mm}
\end{wrapfigure}

%% file: figs/human_eval_app.tex
% \begin{figure}[h!]
% 	\centering
% 	\includegraphics[width=0.7\columnwidth]{figs/human_eval_app.pdf}
% 	\caption{
% \textbf{iOS app for human evaluation}. We developed a simple iOS app that enables human participants to text goal, capture an image using iPhone's back camera, send both to a remote server, and receive the predicted action from our \model policy.}
% 	\label{fig:human_eval_app}
% \end{figure}

\begin{wrapfigure}{R}{0.4\textwidth}
\vspace{-7mm}
  \begin{center}
    \includegraphics[width=0.4\textwidth]{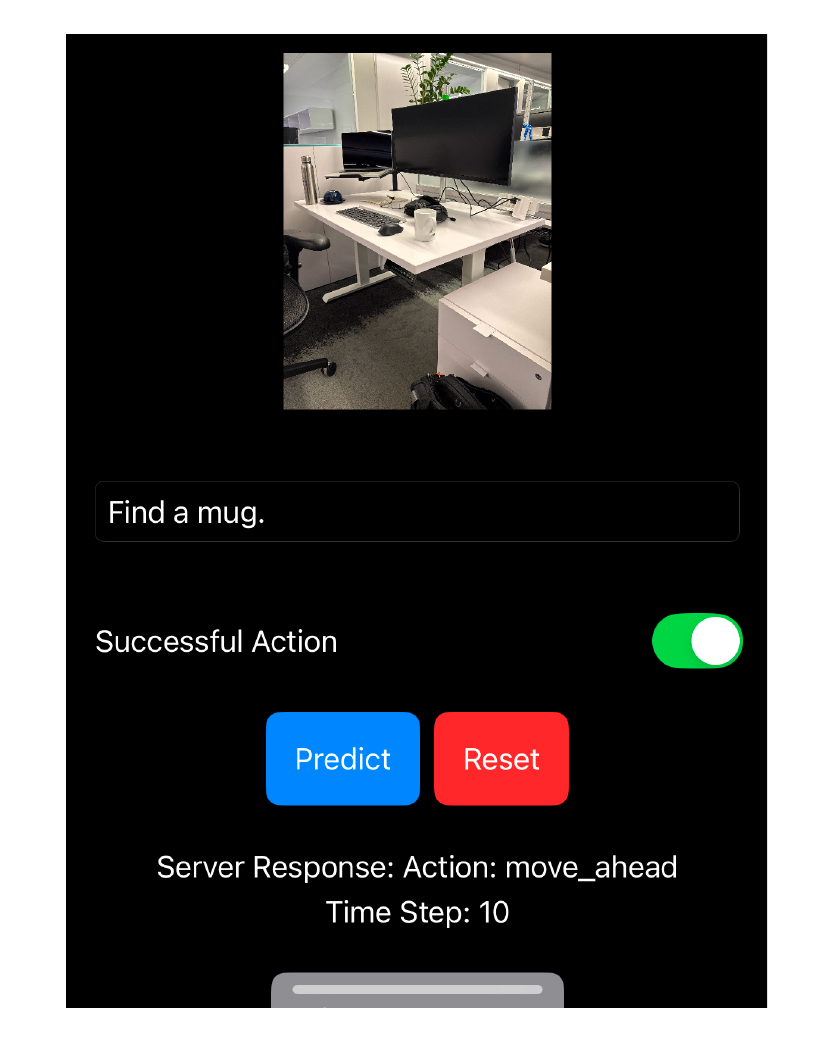}
  \end{center}
  \vspace{-4mm}
  \caption{\footnotesize{textbf{iOS app for human evaluation}. We developed a simple iOS app that enables human participants to text goal, capture an image using iPhone's back camera, send both to a remote server, and receive the predicted action from our \model policy.}}
	\label{fig:human_eval_app}
  \vspace{-4mm}
\end{wrapfigure}

%% file: tables/robot_platforms.tex
% \begin{table*}[!ht]
% \centering
% \small
% \begin{tabular}{lcccc}
% \hline
%  & Stretch RE-1 & Stretch RE-1 \tiny{(Factory)} & LoCoBot & Unitree GO1 \\ \hline
%  Body dimension (cm) & 33 $\times$ 34 $\times$ 141    & 33 $\times$ 34 $\times$ 141 & 35$\times$35$\times$89 & 64.5$\times$28$\times$40       \\ 
%  % \hline
% Camera model      & 2$\times$ D455   & D435  & D435      & D435       \\ 
% % \hline
% % Camera resolution (px)     & $396 \times 224$   & $720 \times 1280$  & $1280 \times 720$      & $1280 \times 720$     \\
% Camera vertical FoV (degrees)     & $59^\circ$  & $69^\circ$  & $42^\circ$      & $42^\circ$      \\
% Camera horizontal FoV (degrees)     & $90^\circ$  & $42^\circ$  & $68^\circ$      & $68^\circ$      \\
% Camera height (cm)     &  140    & 130  & 87      & 28      \\
% % \hline
% Camera pitch (degrees)     & $27^\circ$   & $30^\circ$  & $0^\circ$      & $0^\circ$      \\ 
% \hline
% \end{tabular}
% \caption{\textbf{Details about evaluation robot platforms.} Our four robot platforms have varying dimensions and camera configurations, resulting in diverse evaluation embodiments.}
% \label{tab:robot_spec}
% \end{table*}

\begin{table*}[t]
\centering
\renewcommand{\arraystretch}{1.0}
\scriptsize
\addtolength{\tabcolsep}{0.2em} % TODO added by Jordi - feel free to remove it
\vspace{-2mm}
\begin{tabular}{lcccccc}
\hline
 & Stretch RE-1 & Stretch RE-1 \tiny{(Factory)} & LoCoBot & Unitree GO1 & RB-Y1 (standing) & RB-Y1 (seated)\\ \hline
 Body dimension (cm) & 33$\times$34$\times$141    & 33$\times$34$\times$141 & 35$\times$35$\times$89 & 64.5$\times$28$\times$40 & 60$\times$69$\times$140 &  60$\times$69$\times$92     \\ 
 % \hline
Camera model      & 2$\times$ D455   & D435  & D435      & D435 & iPhone 16 Pro Camera & iPhone 16 Pro Camera       \\ 
% \hline
% Camera resolution (px)     & $396 \times 224$   & $720 \times 1280$  & $1280 \times 720$      & $1280 \times 720$     \\
Camera vertical FoV     & $59^\circ$  & $69^\circ$  & $42^\circ$      & $42^\circ$  & $73^\circ$ & $73^\circ$   \\
Camera horizontal FoV     & $90^\circ$  & $42^\circ$  & $68^\circ$      & $68^\circ$ & $53^\circ$ & $53^\circ$      \\
Camera height (cm)     &  140    & 130  & 87      & 28 & 140 & 92     \\
% \hline
Camera pitch     & $27^\circ$   & $30^\circ$  & $0^\circ$      & $0^\circ$ & $0^\circ$ & $0^\circ$     \\ 
\hline
\end{tabular}
\caption{\footnotesize{\textbf{Details about evaluation robot platforms.} Our four robot platforms have varying dimensions and camera configurations, resulting in diverse evaluation embodiments.}}
\label{tab:robot_spec}
\normalsize
\end{table*}

%% file: tables/human_spec.tex
\setlength{\tabcolsep}{12pt}
\begin{table}[!ht]
\centering
\scriptsize
\begin{tabular}{lccccc}
\hline
 & H1 & H2 & H3 & H4 & H5 \\ \hline
 Height & 6$'$3$''$      & 5$'$10$''$      & 5$'$5$''$      & 6$'$1$''$      & 5$'$11$''$      \\ 
 % \hline
Step size      & 0.25m      & 0.35m      & 0.4m      & 0.3m      & 0.3m      \\ 
% \hline
Rotation Degrees      & 30$^{\circ}$      & 45$^{\circ}$      & 45$^{\circ}$      & 35$^{\circ}$      & 30$^{\circ}$      \\ \hline
\end{tabular}
\vspace{2mm}
\caption{\textbf{Details about human evaluators.} Our five human participants have varying heights, step size, and rotation degrees, resulting in different evaluation embodiments.}
\label{tab:human_spec}
\end{table}
\normalsize

%% file: figs/robot_platforms.tex
\begin{figure}[t]
	\centering
	\includegraphics[width=0.95\columnwidth]{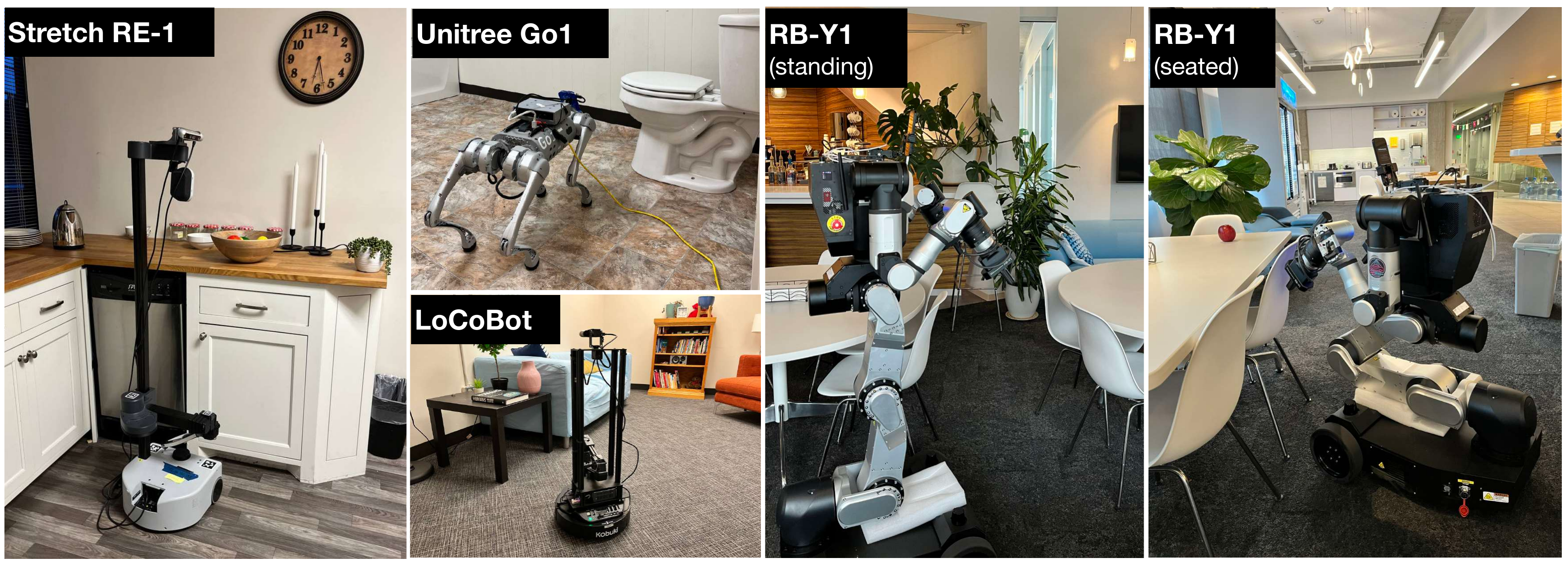}
	\caption{
\footnotesize{\textbf{Robot platforms}. We evaluate on $6$ different platforms: Stretch RE-1 (with two camera configurations), LoCoBot, Unitree GO1, and the RB-Y1 wheeled humanoid, tested in both standing and seated configurations.}}
	\label{fig:robot_platforms}
\end{figure}

% \begin{wrapfigure}{R}{0.6\textwidth}
% \vspace{-7mm}
%   \begin{center}
%     \includegraphics[width=0.6\textwidth]{figs/robot_platforms.pdf}
%   \end{center}
%   \vspace{-4mm}
%   \caption{\footnotesize{\textbf{Robot platforms}. We use $4$ different platforms, including Stretch RE-1 (with two different camera setups), LoCoBot, and Unitree GO1 for our real-world evaluations.}}
% 	\label{fig:robot_platforms}
%   % \vspace{-4mm}
% \end{wrapfigure}

% \begin{table}[t]
% \begin{minipage}{.5\linewidth}
%   \centering
% 	\includegraphics[width=1\columnwidth]{figs/robot_platforms.pdf}
% 	\caption{\footnotesize{\textbf{Robot platforms}. We use $4$ different platforms, including Stretch RE-1 (with two different camera setups), LoCoBot, and Unitree GO1 for our real-world evaluations.}}
% 	\label{fig:robot_platforms}
% \end{minipage}%
% \hspace{0.5cm}
% \begin{minipage}{.5\linewidth}
%   \centering
  
%   \vspace{-2mm}
%   \tiny
%   \setlength{\tabcolsep}{1pt}
%   \renewcommand{\arraystretch}{1.2}
% 	\centering
% 	\includegraphics[width=1\columnwidth]{figs/real_world_layouts.pdf}
% 	\caption{\footnotesize{\textbf{Real Evaluation Environment}. Our real-world evaluations are performed in a multi-room apartment with a long corridor, shown here with the three starting locations for three different robots' evaluations.}}
% 	\label{fig:real_layout}
% \end{minipage}
% % \vspace{-6mm}
% \end{table}

%% file: figs/rainbow_traj.tex
\begin{figure*}
	% \vspace{-1mm}
	%\hspace{4mm}
	\centering
	\includegraphics[width=\textwidth]{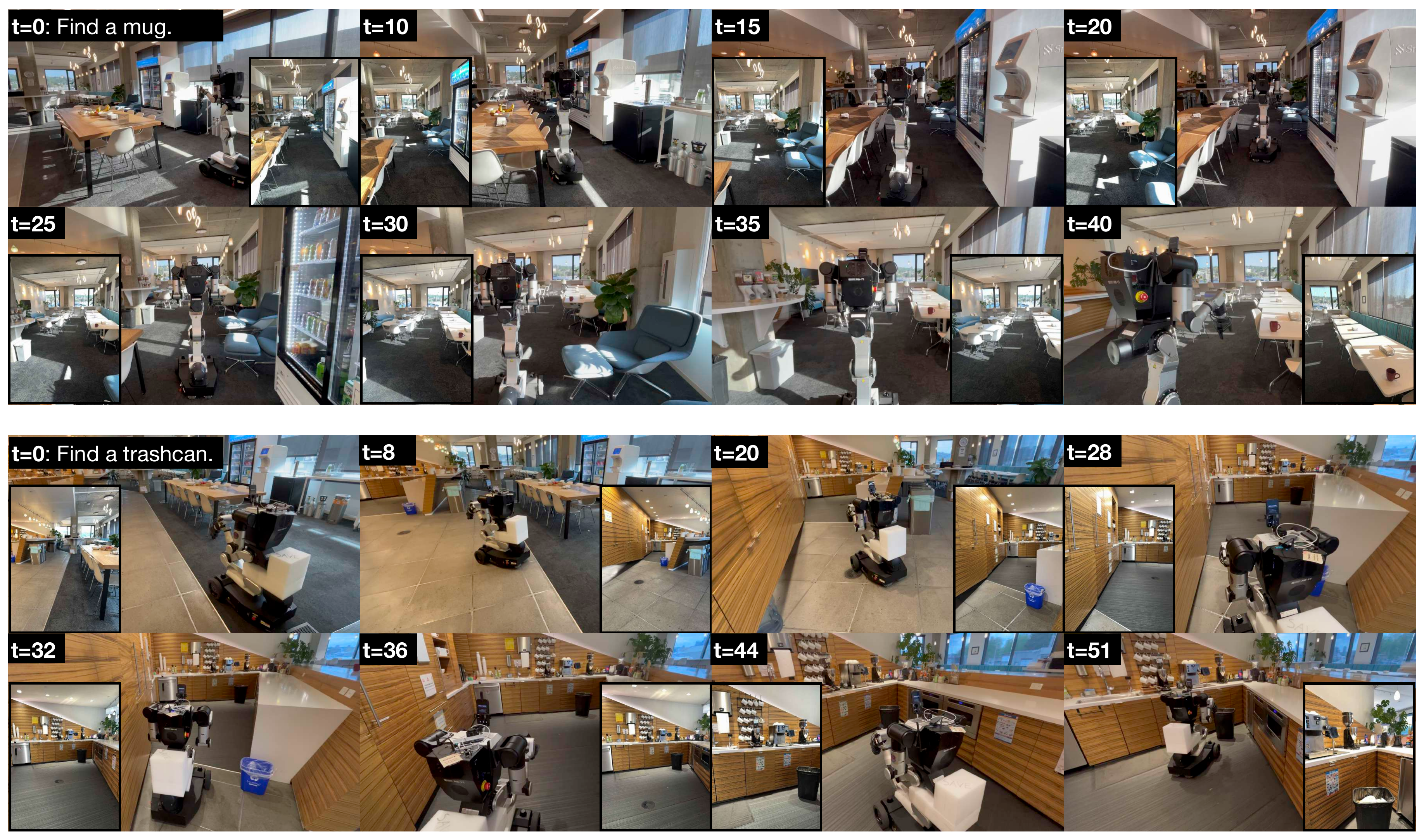}
	\caption{\footnotesize{\textbf{RB-Y1 Trajectories.} We deploy the policy on the wheeled humanoid in an unstructured kitchen area (layout shown in Fig.~\ref{fig:human_eval}) to navigate to different objects. We include both seated and standing configurations and use an iPhone 16 Pro camera to stream the visual observations.
    }}
	\label{fig:rainbow_traj}
\end{figure*}

%% file: figs/human_eval.tex
\begin{figure*}[t]
	%\hspace{4mm}
	\centering
	\includegraphics[width=0.8\columnwidth]{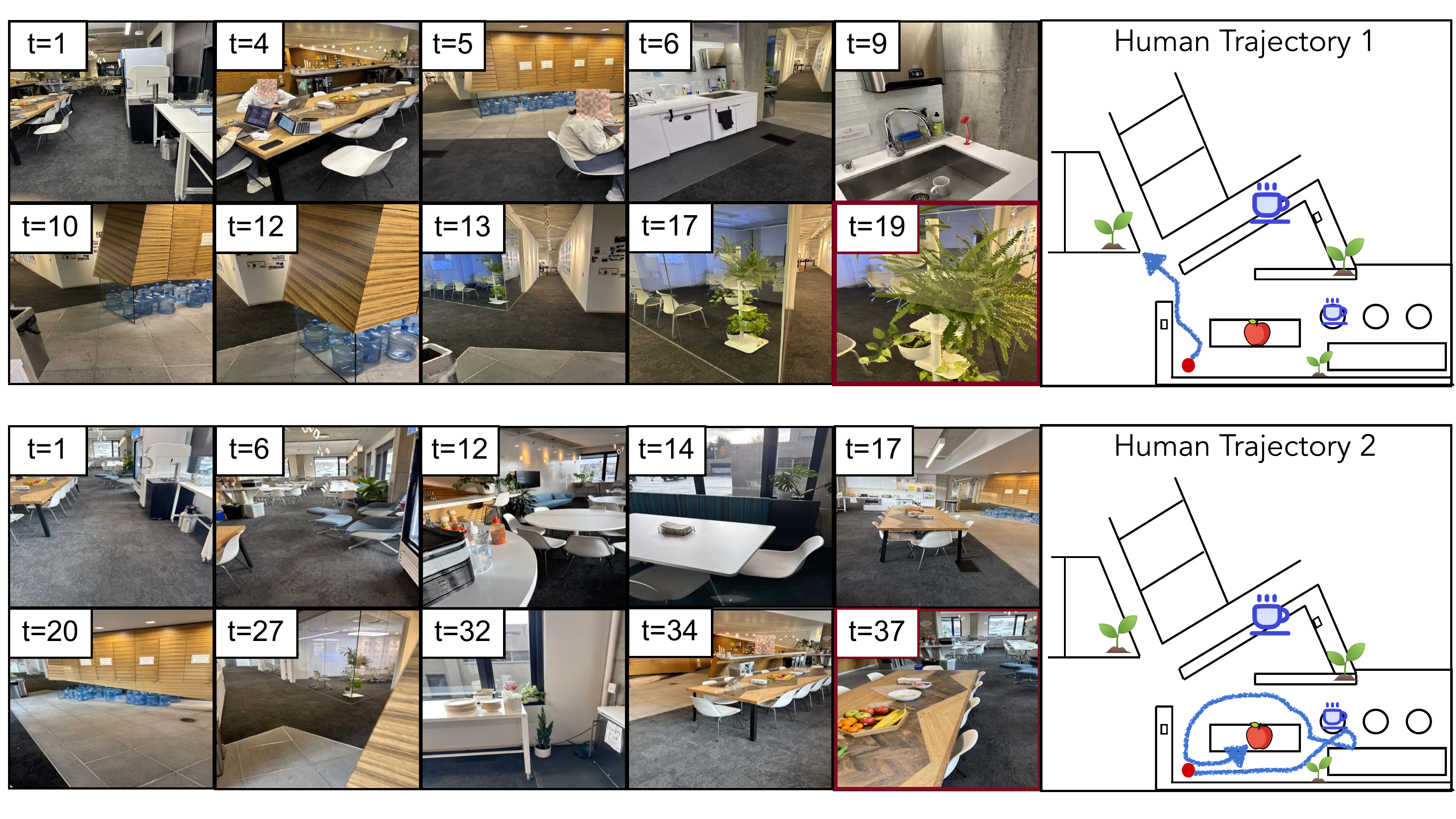}
	\caption{
\textbf{Human Trajectories}. Two sample trajectories from two individuals navigating to a houseplant and an apple using \model.}
	\label{fig:human_eval}
\end{figure*}

% \begin{wrapfigure}{R}{0.5\textwidth}
% \vspace{-5mm}
%   \begin{center}
%     \includegraphics[width=0.5\textwidth]{figs/human_eval.pdf}
%   \end{center}
%   \vspace{-2mm}
%   \caption{\scriptsize{\textbf{Human Trajectories}. Two sample trajectories from two individuals navigating to a houseplant and an apple using \model.}}
% 	\label{fig:human_eval}
%   \vspace{-2mm}
% \end{wrapfigure}

%% file: tables/hyperparams.tex
\begin{table}[h!]
\centering
\small
\begin{tabular}{ll}
\hline 
\multicolumn{2}{c}{ \textbf{Imitation Learning }} \\
\hline Batch Size & 224 \\
Context Length & 100 \\
Learning Rate & 0.0002 \\
\hline 
\multicolumn{2}{c}{ \textbf{RL Finetuning }} \\
\hline Total Rollouts & 64 \\
Learning Rate & 0.0002 \\
Mini Batch per Update & 1 \\
Update Repeats & 4 \\
Max Gradient Norm & 0.5 \\
Discount Value Factor $\gamma$ & 0.99 \\
GAE $\lambda$ & 0.95 \\
PPO Surrogate Objective Clipping & 0.1 \\
Value Loss Weight & 0.5 \\
Entropy Loss Weight & 0.0 \\
Steps for PPO Update & 128 \\
\hline 
\multicolumn{2}{c}{ \textbf{Model Architecture }} \\
\hline 
Transformer State Encoder Layers & 3 \\
Transformer State Encoder Hidden Dims & 512 \\
Transformer State Encoder Heads & 8 \\
Causal Transformer Deocder Layers & 3 \\
Causal Transformer Deocder Hidden Dims & 512 \\
Causal Transformer Deocder Heads & 8 \\
\hline

\end{tabular}
\vspace{2mm}
\caption{\textbf{Hyperparameters for training and model architecture.}}
\label{tab:hyperparams}
\end{table}
\normalsize

%% file: figs/model_details.tex
\begin{figure*}[h!]
    \centering 
    \includegraphics[width=\textwidth]{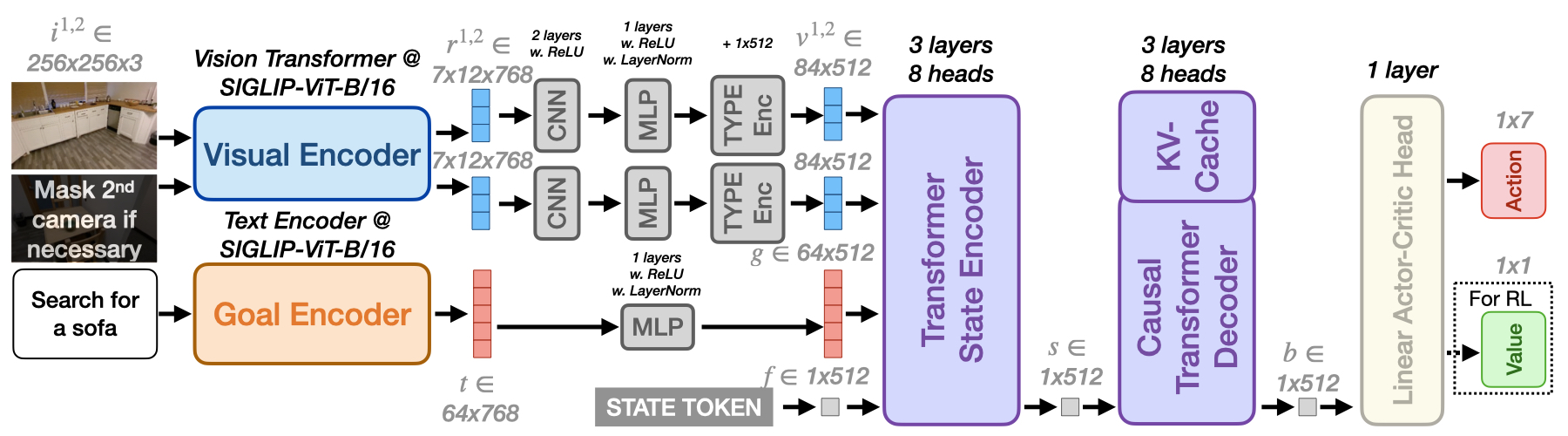}
    \caption{\footnotesize{\textbf{\model architecture}. The notations in gray correspond to hidden feature vectors and the black text on top of each module indicates the hyperparameters for that module.  \model accepts visual observations and a language instruction as inputs and predicts an action to execute. During RL finetuning, \model also predicts a value estimate. We mask the image from the $2^{nd}$ camera with all $0$ for the embodiments with only one camera, such as LoCoBot and Unitree. More specifically, we use the Vision Transformer and the Text Encoder from SIGLIP-ViT-B/16 as our visual encoder and goal encoder. After encoding, we compress and project the visual representation $r$ and text embedding $t$ to $v$ and $g$, respectively, with the desired dimension $d$. Next, the Transformer State Encoder encodes $v$, $g$, along with \texttt{state} token embedding $f$ into a state feature vector $s$. The Causal Transformer Decoder further processes $s$, along with previous experiences stored in the KV-Cache, to produce the state belief $b$. Finally, the Linear Actor-Critic Head predicts action logits (and, during RL finetuning, a value estimate) from $b$.}}
    \label{fig:model_details}
\end{figure*}

%% file: tables/nearest_neighbors_stretch.tex
% \setlength{\tabcolsep}{2pt}
% \begin{table}[t]
% \centering
% \renewcommand{\arraystretch}{1.0}
% \scriptsize
% \begin{tabular}{lllllll}
% & & \multicolumn{5}{c}{Nearest Neighbors} \\
% \cmidrule(r){3-7} 
% & \textbf{Stretch RE-1} & N1 & N2 & N3 & N4 & N5 \\
% \hline 
% Camera Position (x) (meters & 0 & -0.06 & 0.11 & 0 & -0.08 & 0.03 \\
% Camera Position (y) (meters & 1.44 & 1.13 & 0.67 & 0.24 & 0.72 & 0.32 \\
% Camera Position (z) (meters & 0.07 & 0.03 & 0.06 & 0.07 & 0.07 & -0.03 \\
% Camera Pitch (degrees) & 27 & 29 & 33 & 34 & 32 & 33 \\
% Camera Yaw (degrees) & 0 & 0 & 0 & 0 & 0 & 0 \\
% Vertical FoV (degrees) & 59 & 57 & 56 & 54 & 59 & 54 \\
% RGB Resolution (H) & 224 & 224 & 224 & 224 & 224 & 224 \\
% RGB Resolution (Y) & 396 & 394 & 394 & 396 & 396 & 398 \\
% Rotation Center (x) (meters) & 0 & 0 & 0.09 & -0.17 & 0 & 0.02 \\
% Rotation Center (z) (meters) & 0.11 & 0.02 & 0.02 & -0.08 & 0.04 & -0.12 \\
% Collider Size (x) (meters) & 0.34 & 0.23 & 0.28 & 0.49 & 0.33 & 0.24 \\
% Collider Size (y) (meters) & 1.41 & 1.41 & 0.9 & 0.84 & 1.23 & 0.43 \\
% Collider Size (z) (meters) & 0.33 & 0.27 & 0.41 & 0.29 & 0.44 & 0.38 \\
% \hline
% distance & - & 0.38 & 0.7 & 0.79 & 0.8 & 0.92 \\
% \hline
% \end{tabular}
% \normalsize
% \caption{\textbf{Five Nearest Neighbor Embodiments for Stretch RE-1 in Training Data.} }
% \label{tab:nn_stretch}
% % \vspace{-1.5em}
% \end{table}

\begin{table}[t]
\begin{minipage}{.5\linewidth}
  \centering
  
  \vspace{-2mm}
  \tiny
  \setlength{\tabcolsep}{1pt}
  \renewcommand{\arraystretch}{1.2}
\begin{tabular}{lllllll}
& & \multicolumn{5}{c}{Nearest Neighbors} \\
\cmidrule(r){3-7} 
& \textbf{Stretch RE-1} & N1 & N2 & N3 & N4 & N5 \\
\hline 
Camera Position (x) (meters & 0 & -0.06 & 0.11 & 0 & -0.08 & 0.03 \\
Camera Position (y) (meters & 1.44 & 1.13 & 0.67 & 0.24 & 0.72 & 0.32 \\
Camera Position (z) (meters & 0.07 & 0.03 & 0.06 & 0.07 & 0.07 & -0.03 \\
Camera Pitch (degrees) & 27 & 29 & 33 & 34 & 32 & 33 \\
Camera Yaw (degrees) & 0 & 0 & 0 & 0 & 0 & 0 \\
Vertical FoV (degrees) & 59 & 57 & 56 & 54 & 59 & 54 \\
RGB Resolution (H) & 224 & 224 & 224 & 224 & 224 & 224 \\
RGB Resolution (Y) & 396 & 394 & 394 & 396 & 396 & 398 \\
Rotation Center (x) (meters) & 0 & 0 & 0.09 & -0.17 & 0 & 0.02 \\
Rotation Center (z) (meters) & 0.11 & 0.02 & 0.02 & -0.08 & 0.04 & -0.12 \\
Collider Size (x) (meters) & 0.34 & 0.23 & 0.28 & 0.49 & 0.33 & 0.24 \\
Collider Size (y) (meters) & 1.41 & 1.41 & 0.9 & 0.84 & 1.23 & 0.43 \\
Collider Size (z) (meters) & 0.33 & 0.27 & 0.41 & 0.29 & 0.44 & 0.38 \\
\hline
distance & - & 0.38 & 0.7 & 0.79 & 0.8 & 0.92 \\
\hline
\end{tabular}
\caption{\footnotesize{Five Nearest Neighbor Embodiments for Stretch RE-1 in Training Data.} }
\label{tab:nn_stretch}
\end{minipage}%
\hspace{0.5cm}
\begin{minipage}{.45\linewidth}
  \centering
  
  \vspace{-2mm}
  \tiny
  \setlength{\tabcolsep}{1pt}
  \renewcommand{\arraystretch}{1.2}
\begin{tabular}{lllllll}
& & \multicolumn{5}{c}{Nearest Neighbors} \\
\cmidrule(r){3-7} 
& \textbf{Locobot} & N1 & N2 & N3 & N4 & N5 \\
\hline 
Camera Position (x) (meters)& 0 & -0.09 & 0.12 & 0.03 & $-0.06$ & -0.1 \\
Camera Position (y) (meters)& 0.87 & 1.01 & 0.81 & 0.39 & 0.85 & 0.42 \\
Camera Position (z) (meters)& 0 & -0.1 & -0.05 & -0.1 & -0.02 & 0.09 \\
Camera Pitch (degrees) & 0 & 0 & 0 & -1 & 0 & 1 \\
Camera Yaw (degrees) & 0 & 0 & 0 & 0 & 0 & 0 \\
Vertical FoV (degrees)& 42 & 45 & 44 & 42 & 45 & 45 \\
RGB Resolution (H) & 224 & 224 & 224 & 224 & 224 & 224 \\
RGB Resolution (Y) & 396 & 396 & 394 & 394 & 392 & 392 \\
Rotation Center (x) (meters)& 0 & 0.04 & 0.1 & 0.1 & -0.02 & -0.15 \\
Rotation Center (z)  (meters)& 0 & -0.13 & 0 & 0.13 & 0.02 & -0.12 \\
Collider Size (x) (meters)& 0.35 & 0.27 & 0.36 & 0.37 & 0.27 & 0.42 \\
Collider Size (y) (meters)& 0.89 & 1.28 & 1.23 & 0.86 & 1.46 & 0.59 \\
Collider Size (z) (meters)scale z & 0.4 & 0.43 & 0.23 & 0.36 & 0.36 & 0.45 \\
\hline
distance & - & 0.18 & 0.22 & 0.34 & 0.41 & 0.42 \\
\hline
\end{tabular}
\normalsize
\caption{\footnotesize{Five Nearest Neighbor Embodiments for LoCoBot in Training Data.} }
\label{tab:nn_locobot}
\end{minipage}
% \vspace{-6mm}
\end{table}

%% file: tables/nearest_neighbors_unitree.tex
\setlength{\tabcolsep}{2pt}
\begin{table}[t]
\vspace{-2mm}
\centering
\renewcommand{\arraystretch}{1.0}
\scriptsize
\begin{tabular}{lllllll}
& & \multicolumn{5}{c}{Nearest Neighbors} \\
\cmidrule(r){3-7} 
& \textbf{Unitree A1} & N1 & N2 & N3 & N4 & N5 \\
\hline 
Camera Position (x) (meters) & 0.01 & 0.08 & 0.03 & -0.01 & -0.04 & 0.1 \\
Camera Position (y) (meters) & 0.3 & 0.56 & 0.37 & 0.85 & 0.55 & 0.82 \\
Camera Position (z) (meters) & 0.27 & -0.11 & 0.06 & 0 & 0.12 & 0.02 \\
Camera Pitch (degrees) & 0 & -3 & -2 & -4 & -5 & -5 \\
Camera Yaw (degrees) & 0 & 0 & 0 & 0 & 0 & 0 \\
Vertical FoV (degrees) & 42 & 49 & 49 & 51 & 50 & 51 \\
RGB Resolution (H) & 270 & 224 & 224 & 224 & 224 & 224 \\
RGB Resolution (Y) & 480 & 448 & 446 & 448 & 446 & 446 \\
Rotation Center (x) (meters) & 0 & -0.07 & 0.05 & -0.07 & -0.09 & -0.14 \\
Rotation Center (z) (meters) & 0.04 & -0.02 & 0 & -0.12 & 0.12 & 0.11 \\
Collider Size (x) (meters) & 0.3 & 0.46 & 0.27 & 0.35 & 0.27 & 0.49 \\
Collider Size (y) (meters) & 0.34 & 1.24 & 0.45 & 1.47 & 0.67 & 1.39 \\
Collider Size (z) (meters) & 0.64 & 0.34 & 0.37 & 0.36 & 0.33 & 0.39 \\
\hline
distance & - & 0.76 & 0.78 & 1.04 & 1.1 & 1.12 \\
\hline
\end{tabular}
\normalsize
\caption{\footnotesize{Five Nearest Neighbor Embodiments for Unitree A1 in Training Data.} }
\label{tab:nn_unitree}
% \vspace{-1em}
\end{table}

%% file: tables/ablations_vision_encoder.tex
\setlength{\tabcolsep}{10pt}
\begin{table}[t]
\centering
\renewcommand{\arraystretch}{1.1}
% \small
\resizebox{\columnwidth}{!}{
\begin{tabular}{lccccc}
%\toprule
\multirow{2}{*}{Model} & \multirow{2}{*}{Visual Encoder} &  \multicolumn{4}{c}{Benchmark Embodiment} \\
\cmidrule(r){3-6} %\cmidrule(r){4} \cmidrule(r){5}
& & \multicolumn{1}{c}{\textbf{Stretch}} & \multicolumn{1}{c}{\textbf{Stretch} \tiny{(Nav)}} & \multicolumn{1}{c}{\textbf{LoCoBot}} & \multicolumn{1}{c}{\textbf{Unitree A1}} \\
\midrule
\model & \multirow{1}{*}{\textsc{SIGLIP-ViT-B/16}} & 76.0 & 74.0 & 66.5 & 72.0 \\

\modellarge & \multirow{1}{*}{\textsc{ViT-L/14 336px} CLIP} & \textbf{83.8} & \textbf{77.7} & \textbf{75.3} & \textbf{79.9} \\

\midrule

%\bottomrule
\end{tabular}
}
% \normalsize

\caption{\footnotesize{\textbf{A Stronger Visual Encoder}. Using a more powerful vision encoder significantly improves the zero-shot performance across all embodiments.}}
\label{tab:vision_encoder_ablation}
\vspace{-1.0em}
\end{table}

%% file: figs/tsne_nn.tex
\begin{wrapfigure}{R}{0.4\textwidth}
\vspace{-7mm}
  \begin{center}
    \includegraphics[width=0.4\textwidth]{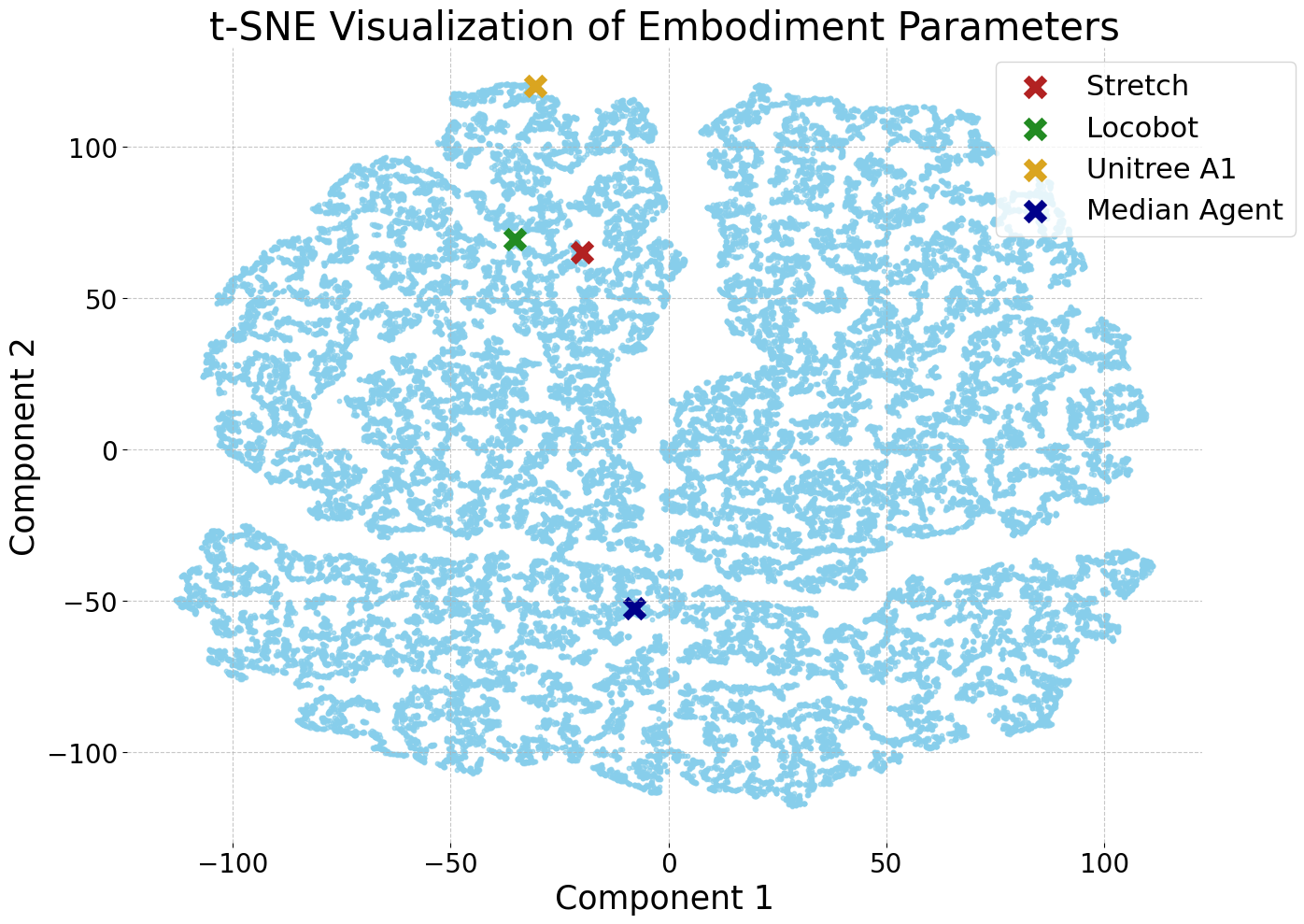}
  \end{center}
  \vspace{-4mm}
  \caption{\footnotesize{t-SNE visualization of the embodiment parameters $\mathbf{c}_e \in \mathbb{R}^{19}$ for 50k random agents. The three specific robots are also shown for visualization (they are not included in our training set).}}
	\label{fig:tsne_nn}
  \vspace{-4mm}
\end{wrapfigure}

%% file: figs/random_embodiments.tex
\begin{figure}[t]
	% \vspace{-1mm}
	%\hspace{4mm}
	\centering
	\includegraphics[width=\textwidth]{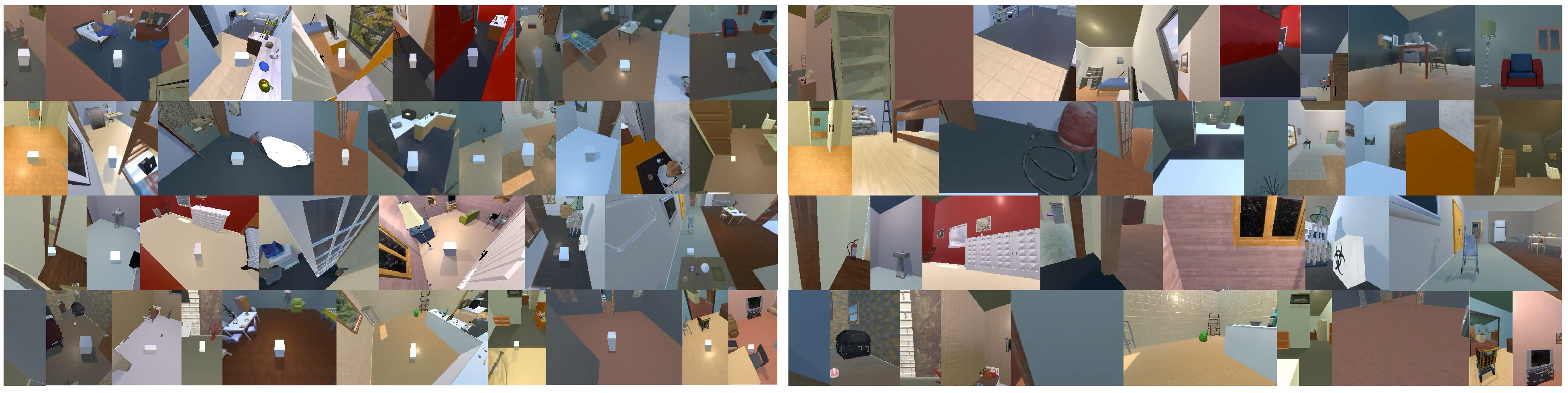}
	\caption{\footnotesize{\textbf{Random embodiments in the AI2-THOR simulator}. Right column shows the egocentric view from the main camera and the left column shows a third-person view of the agent --white boxes indicate the robot colliders for visualization purposes only.
    }}
	\label{fig:random_embodiments}
\end{figure}

%% file: figs/ood_ranges.tex
\begin{figure}
	%\hspace{4mm}
	\centering
	\includegraphics[width=0.9\columnwidth]{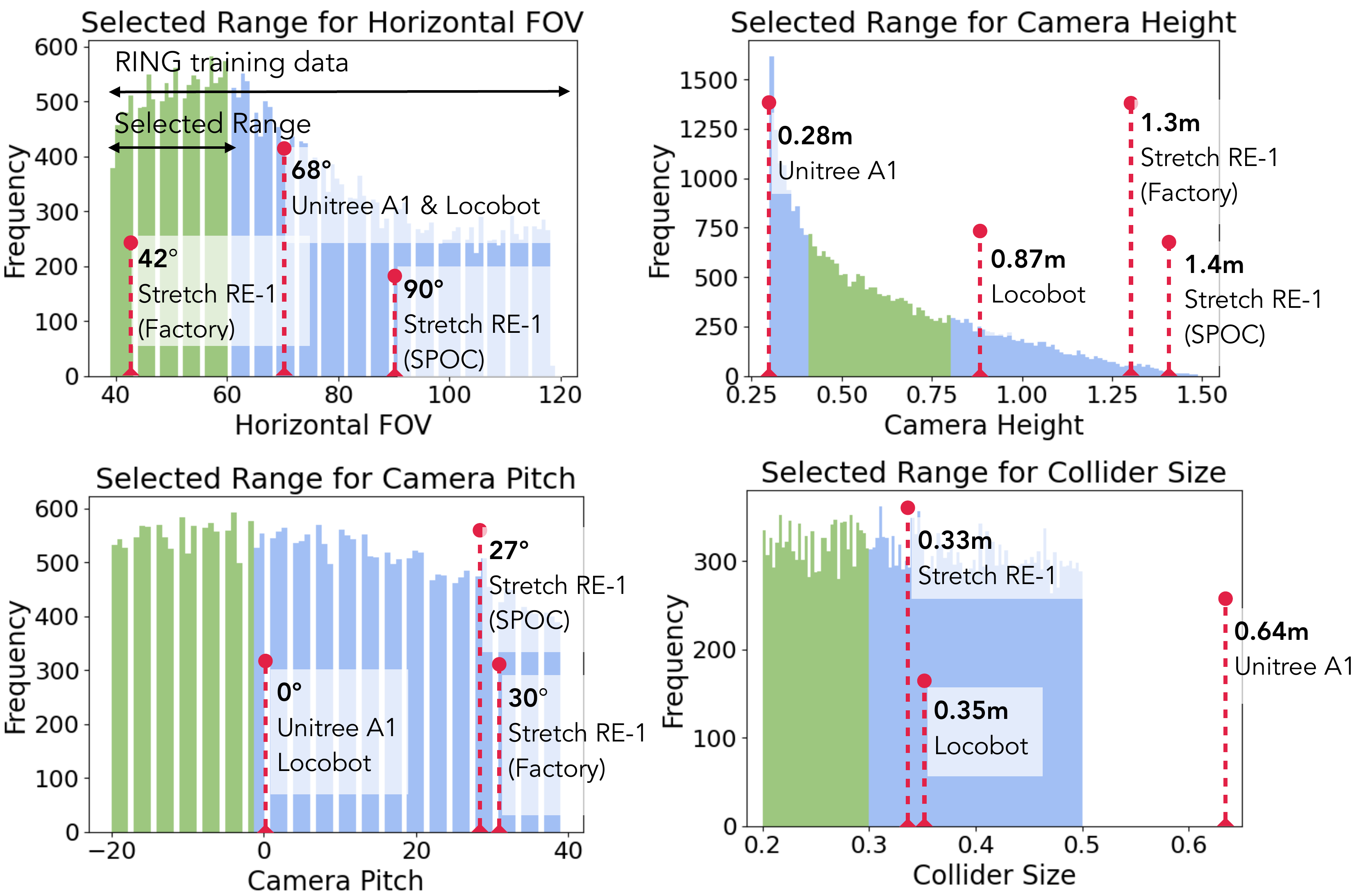}
	\caption{
\footnotesize{\textbf{Selected training ranges for the four embodiment parameters (camera height, camera FOV, camera pitch, and collider size)}. The green regions represent the narrower ranges used during training, excluding the values corresponding to the real robots. The results of policies trained on each selected range are presented in Table~\ref{tab:ood_results}.}}
	\label{fig:ood_ranges}
\end{figure}

%% file: tables/ood_generalization.tex
\begin{table*}[t]
\centering
\renewcommand{\arraystretch}{1.2}
\scriptsize
\addtolength{\tabcolsep}{0.05em} % Adjusts column spacing
\begin{tabular}{lllllll}
\multirow{2}{*}{Embodiment} & \multirow{2}{*}{Training Range} & \multicolumn{5}{c}{Success Rate$\uparrow$ / Collision Rate $\downarrow$} \\
\cmidrule(r){3-7} 
Parameter & & \textbf{Stretch} & \textbf{Stretch} \tiny{(Factory Config)} & \textbf{LoCoBot} & \textbf{Unitree A1} & \textbf{Average} \\
\midrule
Camera Height & [0.4, 0.8] & 55.3 / \textbf{9.3} & 51.0 / 9.8 & 51.5 / 9.8 & 59.3 / 9.2 & 54.3 / 9.5 \\
Camera FoV & [40, 60] & 54.0 / 14.3 & 51.6 / 15.5 & 53.4 / 12.8 & \textbf{61.5} / 11.5 & 55.1 / 13.5 \\
Camera Pitch & [-20, -2] & 54.5 / 12.9 & 53.5 / 9.7 & \textbf{56.5} / 11.2 & 59.8 / 12.7 & 56.1 / 11.6 \\
Collider Size & [0.20, 0.32] & \textbf{60.5} / 18.0 & 53.5 / 21.4 & 55.0 / 14.9 & 54.0 / 18.6 & 55.7 / 18.2 \\
\bottomrule
No Filter & - & 58.8 / 9.6 & \textbf{60.0} / \textbf{9.5} & \textbf{56.5} / \textbf{7.9} & 60.9 / \textbf{8.3} & \textbf{59.1} / \textbf{8.8}  \\
\end{tabular}
\normalsize

\caption{\footnotesize{\textbf{Out-of-distribution (OOD) generalization for different embodiment parameters.} For each of the four embodiment parameters, we select 50k random embodiments from a narrow training range that excludes the parameter values of the real robots (as shown in Fig.~\ref{fig:ood_ranges}). Zero-shot evaluations are performed on four real robots, each with parameter values outside the training distribution. Success and collision rates are reported for each robot and averaged across all robots.}}
\label{tab:ood_results}
% \vspace{-1.5em}
\end{table*}